%% file: main.tex
\documentclass[%
 reprint,
 amsmath,amssymb,
 aps,
]{revtex4-2}

\usepackage{amsmath,amssymb,amsfonts,mathrsfs,mathtools}
\usepackage{ dsfont }
\usepackage{xcolor}
\usepackage{graphicx}% Include figure files
\usepackage{dcolumn}% Align table columns on decimal point
\usepackage{bm}% bold math
\usepackage[separate-uncertainty = true,multi-part-units=single]{siunitx}
\usepackage{natbib}
\usepackage{booktabs}
\usepackage{makecell}
\usepackage{float}

\usepackage[hidelinks]{hyperref}
\hypersetup{
    colorlinks=true,
    linkcolor=blue,   % all internal links (including cref) black
    citecolor=blue,    % optional: color for citations
    urlcolor=blue      % optional: color for URLs
}
\usepackage[capitalise,nameinlink]{cleveref}
\usepackage[normalem]{ulem}
\usepackage{svg}
\usepackage[mathlines]{lineno}
\begin{document}

\makeatletter
\renewcommand\section{\@startsection{section}{1}{0pt}%
  {-2.0ex plus -0.5ex minus -0.2ex}%
  {1.0ex plus 0.2ex}%
  {\normalfont\large\bfseries}}
\renewcommand\subsection{\@startsection{subsection}{2}{0pt}%
  {-1.5ex plus -0.5ex minus -0.2ex}%
  {0.5ex plus 0.1ex}%
  {\normalfont\normalsize\bfseries}}
\makeatother

\newcommand{\kai}[1]{\textcolor{orange}{K: #1}}
\newcommand{\luis}[1]{\textcolor{blue}{L: #1}}
\newcommand{\rachel}[1]{\textcolor{green}{R: #1}}
\newcommand{\niao}[1]{\textcolor{yellow}{N: #1}}
\newcommand{\todo}[1]{\textcolor{blue}{TODO: #1}}
\newcommand{\red}[1]{\textcolor{red}{#1}}
\preprint{APS/123-QED}

\title{Integrated electro-optic attention \\nonlinearities for transformers}

\author{Luis Mickeler\textsuperscript{1}\thanks{Corresponding author: lmickeler@phys.ethz.ch}, 
Kai Lion\textsuperscript{2}, 
Alfonso Nardi\textsuperscript{3},
Jost Kellner\textsuperscript{1}, 
Pierre Didier\textsuperscript{1},\\
Bhavin J. Shastri\textsuperscript{4},
Niao He\textsuperscript{2}, 
Rachel Grange\textsuperscript{1}}

\affiliation{\textsuperscript{1}Optical Nanomaterial Group, Department of Physics, ETH Zurich, Zurich, Switzerland}

\affiliation{\textsuperscript{2}Optimization \& Decision Intelligence Group, Department of Computer Science, ETH Zurich, Zurich, Switzerland}

\affiliation{\textsuperscript{3}Department of Physics, Politecnico di Milano, Milan, Italy}

\affiliation{\textsuperscript{4}Centre for Nanophotonics, Department of Physics, Engineering Physics \& Astronomy, Queen’s University, Kingston,  Canada}

\begin{abstract}
Transformers achieve state-of-the-art performance in language processing and computer vision. Their computation increasingly depends on hardware that can keep pace not only with large matrix multiplications, but with nonlinear operations like Softmax within the core attention mechanism. On digital accelerators, Softmax accounts for only a small fraction of total operations, yet it can disproportionately impact overall inference latency. Here, we use thin-film lithium niobate Mach-Zehnder modulators (MZM) as analog computing elements for low-latency Transformer attention. Our high-speed nonlinearities based on the intrinsic sinusoidal MZM response achieve performance close to digital baselines in Vision Transformers and GPT-2, even under 4-bit input and output quantization.
We experimentally characterize the electro-optic response at symbol rates up to \SI{10}{GBaud} and assess model robustness under corresponding noise conditions. Accounting for the full signal chain, our analysis indicates more than an order of magnitude lower latency relative to previous custom electronic and photonic Softmax implementations. These results establish analog electro-optic devices as nonlinear function units well suited for integration into emerging co-packaged electronic–photonic hardware, offering a path toward high-speed transformer inference.
\end{abstract}

\maketitle

\input{01_introduction}

\section*{Results}
\input{03_results}

\section*{Discussion}
\input{04_discussion}

\section*{Methods}
\input{05_methods}

\newpage
\bibliographystyle{naturemag}
\setcitestyle{numbers,square}
\bibliography{refs}

\section*{Author contributions}
L.M. and K.L. performed transformer trainings. L.M., J.K. and P.D. conducted experimental measurements. A.N. and L.M. performed the energy and latency extrapolations. L.M. and K.L. conceived the original ideas. B.S., N.H. and R.G. supervised the project. N.H. and R.G. acquired funding. All authors contributed to the discussion and paper revision.

\section*{Competing interest}
The authors declare no competing financial or nonfinancial interests.

\begin{acknowledgments}
This work was supported by the Swiss National Science Foundation SNSF Consolidator Grant APIC (TMCG-2\_213713), by Sinergia LION (CRII5-216600), and by the European Union's Horizon Europe research and innovation programme under the Marie Skłodowska-Curie Actions HORIZON-MSCA-2024-PF-01-01 grant agreement No. 101203711 (A.N.). We thank Alessandra Sabatti for the fabrication of the measured modulators. We acknowledge support for the fabrication from the cleanroom facilities BRNC and FIRST of ETH Zurich and IBM Rüschlikon. 
\end{acknowledgments}

\section*{Data availability statement}
The code used to generate the results of this study is publicly available from \url{https://github.com/LuisMickeler/optical-attention-nonlin}. The connected data is publicly available from \url{http://hdl.handle.net/20.500.11850/805674}.\\

\newpage

\clearpage
\appendix
\renewcommand{\thesection}{S\arabic{section}}
\input{xx_supplementary}

\end{document}

%% file: 01_introduction.tex
\begin{figure*}[t]
    \centering    
    \includegraphics[width=1\linewidth]{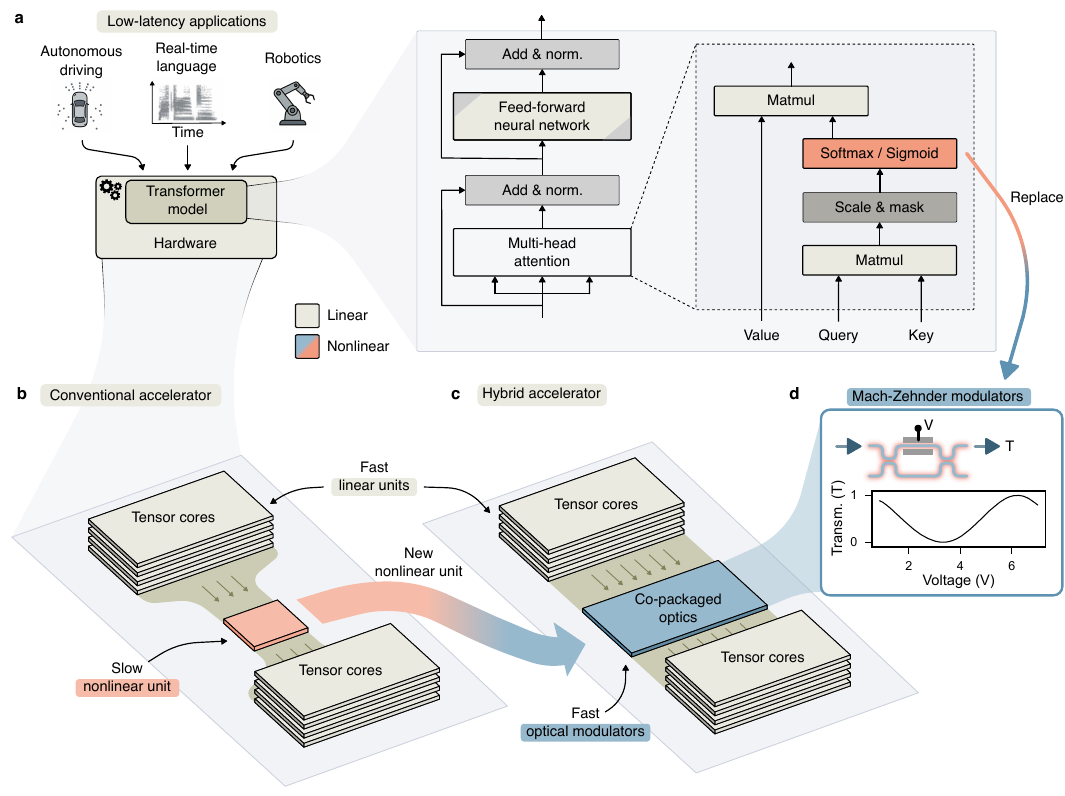}
    \caption{\textbf{Accelerating attention nonlinearities in transformer with co-packaged optics} \textbf{(a)} Latency-critical applications using transformer models, where each Transformer block consists of multi-head attention and feed-forward layers \cite{vaswani_attention_2017}. The self-attention computation relies on a nonlinear reweighting using Softmax/Sigmoid nonlinearity (targeted in this work). Matmul refers to matrix multiplication. Mask refers to the application of a triangular mask to prevent attending to future tokens in the sequence. \textbf{(b)} Conventional digital accelerators rely on fast tensor cores for matrix multiplications, but scarce low-throughput special-function units (orange) set a latency bottleneck. \textbf{(c)} A hybrid accelerator leveraging co-packaged optics as active nonlinear computational elements to accelerate attention. \textbf{(d)} The Mach-Zehnder modulators in co-packaged optics exhibit a nonlinear sinusoidal optical response to the supplied voltage.}
    \label{fig:attention-figure}
\end{figure*}

Transformer models have achieved state-of-the-art performance across natural language processing and computer vision tasks \cite{vaswani_attention_2017, guo_deepseek-r1_2025, dosovitskiy_vit_2021}. They process inputs through layers that combine attention—which captures contextual dependencies among input elements (e.g., subword tokens in a sentence or regions in an image)—with fully connected feed-forward layers. Within each attention layer, the input is first linearly projected into the \textit{Query}, \textit{Key}, and \textit{Value} matrices. Matrix multiplication of the queries and keys yields attention scores that quantify pairwise relationships across the input. A subsequent attention nonlinearity converts these unbounded scores into probability-like weights. Typically, the Softmax function performs this non-linearity \cite{vaswani_attention_2017, bai_survey_2026}, with recent architectures also adopting Sigmoid-based variants \cite{ramapuram_sigmoid_2025}.

The remarkable capabilities of transformers have largely emerged through scaling: increases in computing resources, model size, and training data have consistently improved performance across a broad range of applications \cite{NEURIPS2022_c1e2faff, bai_survey_2026}. As a result, modern transformer models now routinely contain billions to trillions of parameters \cite{kimiteam2026kimik2openagentic, team_kimi_2026}, imposing substantial costs in inference latency, energy consumption, and hardware complexity \cite{NEURIPS2022_c1e2faff, kaplan2020scalinglawsneurallanguage}. These constraints are particularly critical in time-sensitive applications such as real-time vision \cite{1Carion_endtoend, pmlr-v139-bertasius21a}, conversational systems \cite{henderson-etal-2020-convert}, and robotic control  \cite{decision_transformer, rt12022arxiv}.

Analog photonics, which uses light for computation, has emerged as a promising platform for ultra-low-latency neural-network inference, with implementations spanning both integrated \cite{feldmann_parallel_2021, shen_deep_2017, ahmed_universal_2025, hua_integrated_2025, al-kayed_programmable_2025} and free-space approaches \cite{lin_all-optical_2018, zhou_large-scale_2021, Spall:20, PhysRevX.9.021032}. Most reported systems accelerate general matrix-matrix multiplications (GEMMs), the linear operations that dominate neural network computation \cite{mcmahon_physics_2023}. Yet, to this day, the matrix dimensions of modern trillion-parameter transformers exceed the scaling limits of current photonic accelerators \cite{bruckerhoff-pluckelmann_case_2026, brunner2025roadmapneuromorphicphotonics}, limiting their practical use for these large linear operations.

In transformer inference, however, latency depends not only on large GEMMs but on the far fewer nonlinear Softmax computations in attention. This shifts the latency bottleneck from many linear operations to few nonlinear ones \cite{Stevens_Softermax}. At a sequence length of $n=8192$, for example, GEMMs account for \SI{99.43}{\%} of total operations in GPT-2, and Softmax for only \SI{0.56}{\percent} \cite{radford_language_2019, hoffmann2022training, kaplan2020scalinglawsneurallanguage}. Nevertheless, Softmax can consume \SI{22}{\percent} of the total inference time (see \cref{supp:softmax_bottleneck}). This disparity stems from graphics processing unit (GPU) architecture: highly parallel tensor cores execute linear matrix multiplications at substantially higher throughput than nonlinear units execute exponential functions required by Softmax. Transformers amplify this disparity by applying Softmax throughout the attention layers. By contrast, conventional neural networks often use it only in the final classification layer.

Efforts to address this bottleneck span both software and hardware \cite{bai_survey_2026, wang_vexp_2025, karami_understanding_2025, wang_sole_2023, xia_hyft_2024, li_high_2023, kim_full_2023}. Software approaches include the \textsc{FlashAttention} series, which combines the steps of attention into a single optimized GPU routine to reduce data movement and execution overhead \cite{dao2022flashattention, dao2023flashattention2, shah2024flashattention, zadouri2026flashattention4algorithmkernelpipelining}. Its latest iteration directly addresses the nonlinear-throughput disparity of GPUs by approximating the exponential function using linear integer arithmetic \cite{zadouri2026flashattention4algorithmkernelpipelining}. Other approaches approximate the exponential using Padé functions, polynomials, linear functions, or neural networks \cite{sadeghi_peano-vit_2024, koca_hardware-efficient_2023,yu_nn-lut_2022, kim2021bert}. Hardware approaches include Softmax-specific digital circuits \cite{Stevens_Softermax}, dedicated electronic units for computing exponentials \cite{wang_vexp_2025}, and analog electronic architectures \cite{leroux_analog_2025, sillman2023analogimplementationsoftmaxfunction, 1010938}. Photonic Softmax has also been demonstrated in free-space electro-optic systems \cite{zhan_optoelectronic_2024} and integrated devices for all-optical computation \cite{dash_softonic_2025, park2026photonicexponentialapproximationcascaded}. However, these approaches rely on bulky free-space optics, optical lookup tables, repeated electro–optic conversions, or precisely tuned microring resonators, which constrain scalability and add latency. 

Here, we instead exploit the intrinsic nonlinear response of a Mach-Zehnder modulator (MZM) to approximate the Softmax and Sigmoid functions used in transformer attention (\cref{fig:attention-figure}\textcolor{blue}{a}). An MZM converts an electrical input into a controlled change in optical intensity. In co-packaged optics, which integrates electronic and photonic components within the same hardware package, MZMs are conventionally operated in a near-linear regime for high-bandwidth communication \cite{10876612, kim_co-packaged_2026}. We instead use their nonlinear response for analog computation. To address the throughput disparity between linear and nonlinear operations (\cref{fig:attention-figure}\textcolor{blue}{b}), we adopt a hybrid approach that pairs digital GEMMs on GPUs with high-speed analog nonlinearities (\cref{fig:attention-figure}\textcolor{blue}{c-d}). This eliminates the need for optical amplification, photonic lookup tables, and deeply cascaded electro–optic conversions. We characterize the MZM-based nonlinearities (which we term \textit{Optmax} and \textit{Optmoid}) at symbol rates up to $\SI{10}{GBaud}$ and evaluate them as direct replacements for digital Softmax and Sigmoid in a Vision Transformer and a Generative Pre-Trained Transformer (GPT). Across a range of quantization and noise levels, our approach achieves performance comparable to digital Softmax and Sigmoid baselines, while reducing the latency of attention nonlinearity by more than one order of magnitude relative to prior custom electronic and photonic Softmax implementations. These results extend beyond MZMs to other physical nonlinearities and position co-packaged optics as a promising platform for nonlinear low-latency computing in large-scale transformers.

%% file: 03_results.tex
\label{results}

\begin{figure*}[t]
    \centering
    \includegraphics[width=1\linewidth]{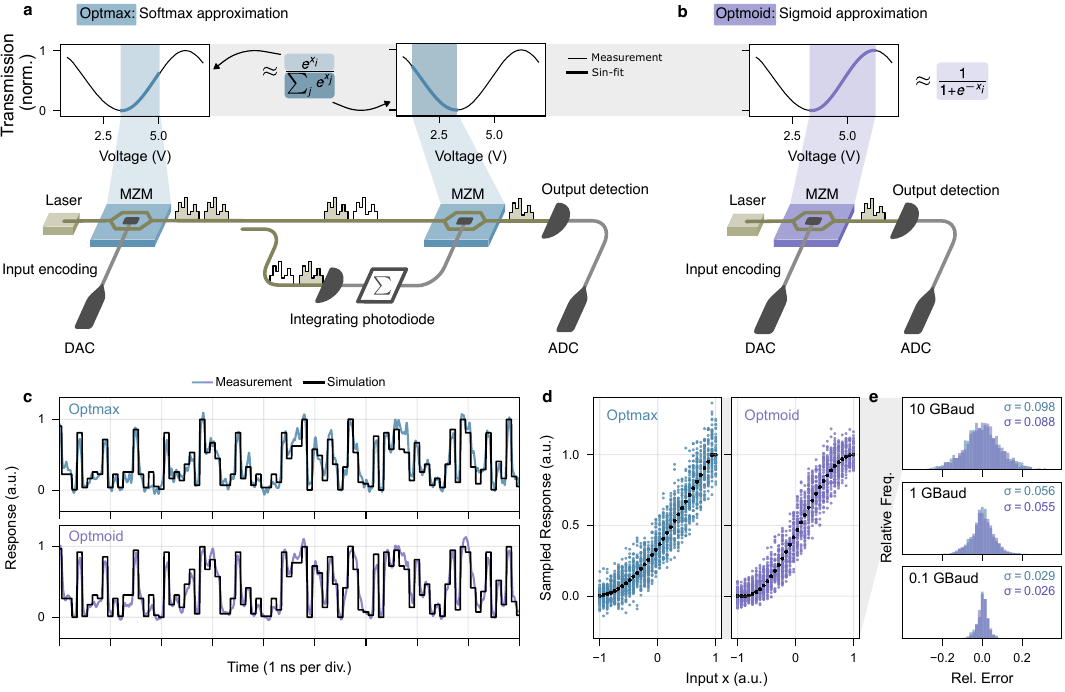}
    \caption{\textbf{Architecture and experimental validation of the \textit{Optmax} and \textit{Optmoid} attention mechanisms}. \textbf{(a)} The \textit{Optmax} architecture approximating the Softmax function: the rising sinusoidal slope of a first MZM mimics the exponential numerator, while the falling slope of a second MZM approximates the reciprocal normalization. The time-multiplexed optical intensities are integrated on a low-bandwidth photodiode to modulate the second MZM before high-speed detection. \textbf{(b)} The \textit{Optmoid} architecture approximates the Sigmoid function using the full min-to-max swing of a single MZM, thereby eliminating the reciprocal-normalization arm required by \textit{Optmax}. \textbf{(c)} A comparison between the simulated (black) and measured (colored) responses for a sequence at 5-bit resolution and $\SI{10}{GBaud}$. \textbf{(d)} The transfer function mapping digital input values to sampled optical amplitudes at $\SI{10}{GBaud}$ for \textit{Optmax} and \textit{Optmoid}; black markers denote the target response. \textbf{(e)} The statistical distribution of the relative error for both nonlinearities across symbol rates of 100 MBaud, 1 GBaud, and 10 GBaud.}
    \label{fig:2}
\end{figure*}
\subsection*{Electro-optic attention nonlinearities}

\textit{Optmax} realizes a Softmax drop-in replacement by mapping exponentiation and normalization onto the rising and falling slopes of two cascaded MZMs. \cref{fig:2}\textcolor{blue}{a} illustrates the signal chain: digital input values $x_i$ are first clipped to the set range $[x_{\min}, x_{\max}]$ and converted to analog voltages via a high-speed digital-to-analog converter (DAC). These input voltages drive the first MZM, modulating a continuous-wave laser source. By biasing the MZM to operate along the rising edge, the resulting optical power approximates the exponential terms. To approximate the Softmax denominator $z = \sum e^{x_j}$, the time-multiplexed optical intensities are encoded twice. The first train of intensities is directed via a tunable coupler to a low-bandwidth photodiode (PD), which integrates the total optical power over the sequence length. The resulting voltage drives a second MZM, which modulates the second train. Since this second MZM is constrained to operate in its falling slope, the modulation approximates the multiplication by the reciprocal $1/z$. The final signals are detected by a high-speed photodiode and digitized by an analog-to-digital converter (ADC).

A fundamental distinction between \textit{Optmax} and the standard Softmax function arises from the physical constraints of the MZM’s sinusoidal response: a modulator's transmission is bounded by unity. The signal-to-noise ratio of the analog computing system additionally limits the lower bound. We note that this sets a limitation on how well both the unbounded exponential and reciprocal of the Softmax can be approximated.

Using Sigmoid as an attention nonlinearity sidesteps this limitation, since the output is intrinsically confined between zero and one~\cite{ramapuram_sigmoid_2025, yan2025sigmoidselfattentionlowersample}. While Sigmoid still necessitates element-wise exponential and reciprocal computation, it does not perform a sum-wise normalization. The sinusoidal MZM response is thus a natural fit. Our proposed \textit{Optmoid} approximates the Sigmoid with the full min-to-max swing of an MZM response function. The core device, shown in \cref{fig:2}\textcolor{blue}{b}, is thus embodied in just a single MZM, alleviating the need for a second modulator or optical arm.

We measure both nonlinearities by driving fabricated thin-film lithium niobate (TFLN) MZMs at symbol rates of up to $\SI{10}{GBaud}$ (see Methods) and compare their output with the differentiable model used in simulation. The measured responses of random \SI{5}{bit} input sequences of length $n=2048$ track the simulations well (\cref{fig:2}\textcolor{blue}{c}). We quantify simulation-to-experimental deviations as amplitude noise by analyzing input values against corresponding sampled optical amplitudes (\cref{fig:2}\textcolor{blue}{d}). A histogram of the relative error (\cref{fig:2}\textcolor{blue}{e}) returns $\sigma \approx 0.1$ at \SI{10}{GBaud} for both nonlinearities, falling to $\sigma \approx 0.03$ at \SI{100}{MBaud}. This measured noise floor sets the operating point at which we probe noise robustness in the following sections.

We insert \textit{Optmax} and \textit{Optmoid} as drop-in replacements for Softmax and Sigmoid within the attention module of otherwise standard transformers, using their differentiable fits for backpropagation. The finite DAC/ADC resolution is accounted for by quantizing the nonlinearity input and output to a chosen bit-width. Output quantization is applied to the Softmax/Sigmoid baselines for a fair comparison, and noise is injected at the nonlinearity output during inference to reproduce the measured analog fluctuations. While Softmax requires no external parameterization, the clipping ranges for \textit{Optmax} and the biases for Sigmoid and \textit{Optmoid} are calibrated prior to each training run (see Methods). We evaluate the resulting models on vision and language benchmarks below.

\subsection*{Image Classification}
\label{vit_results}

\begin{figure*}[t]
    \centering
    \includegraphics[width=1\linewidth]{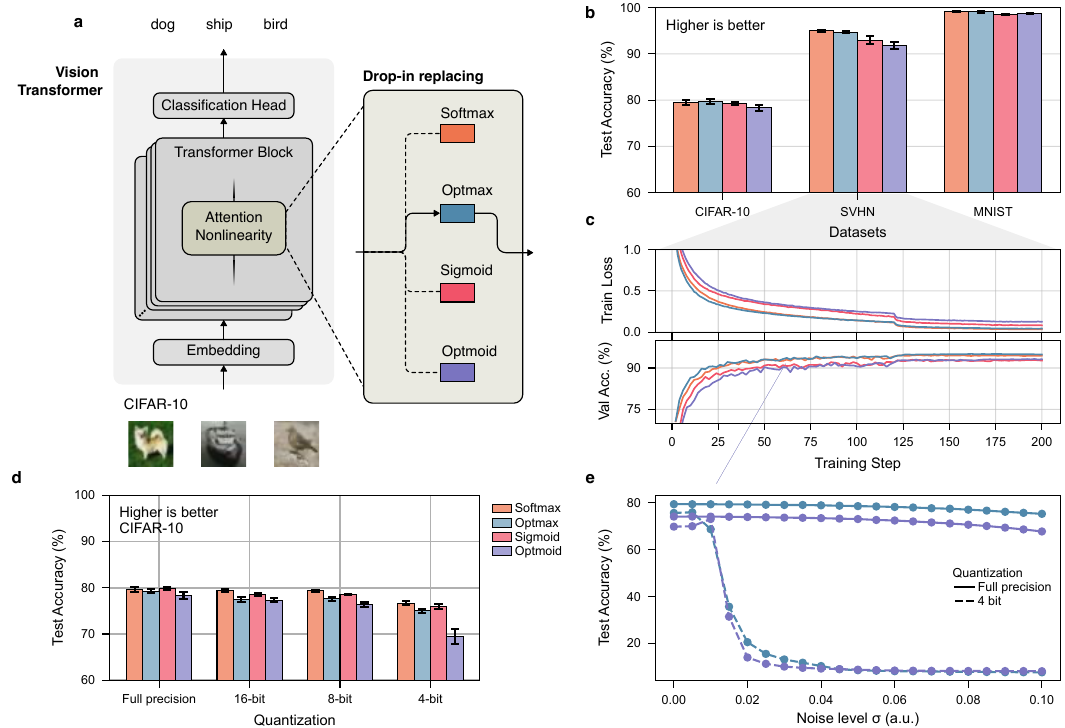}
    \caption{\textbf{Vision Transformer (ViT) performance using electro-optic nonlinearities.} \textbf{(a)} A vision transformer classifies images using stacked transformer blocks. We drop-in replace different attention nonlinearities.
    \textbf{(b)} Final test accuracy comparison for ViT models on MNIST, CIFAR-10, and SVHN datasets. The proposed \textit{Optmax} and \textit{Optmoid} demonstrate competitive performance relative to standard Softmax and Sigmoid benchmarks. 
    \textbf{(c)} Training loss and validation accuracy curves for the CIFAR-10 dataset, illustrating stable convergence across all tested activation functions. 
    \textbf{(d)} Quantization sensitivity analysis showing test accuracy as a function of bit-depth (full precision down to 4-bit).
    \textbf{(e)} Noise robustness showing final test accuracy under additive Gaussian noise $\mathcal{N}(0, \sigma^2)$ (no noise during training).}
    \label{fig:vit_results}
\end{figure*}

As a first proof of concept, we benchmark both nonlinearities in a Vision Transformer \cite{dosovitskiy_vit_2021} (ViT) on the MNIST, CIFAR-10, and SVHN datasets \citep{mnist_2010, cifar_2009, svhn}. The ViT processes images by partitioning them into patches, which are subsequently flattened and mapped to an embedding space. After the addition of positional embeddings, the resulting sequence is processed by a stack of transformer blocks in which we drop-in replace Softmax (\cref{fig:vit_results}\textcolor{blue}{a}). To ensure statistical robustness, we report the accuracy on a held-out test set, averaged across three independent runs with randomized initializations and batch orderings. 

Our proposed nonlinearities (\textit{Optmoid} and \textit{Optmax}) remain competitive with Sigmoid and Softmax baselines, across most benchmarks (\cref{fig:vit_results}\textcolor{blue}{b}). On the CIFAR-10 dataset, \textit{Optmax} achieves a test-accuracy of $\SI{79.73 \pm 0.55}{\percent}$ and thus matches (within error-bars) the performance of Softmax and Sigmoid baselines ($\SI{79.49\pm0.56}{\percent}$ and $\SI{79.32\pm0.38}{\percent}$). \textit{Optmoid} achieves a final test-accuracy of $\SI{78.32\pm0.58}{\percent}$, barely below the Sigmoid. On both SVHN and MNIST datasets, \textit{Optmax} and \textit{Optmoid} match the performance of Softmax and Sigmoid, respectively, within error-bars. \cref{fig:vit_results}\textcolor{blue}{c} depicts the corresponding validation accuracy and training loss curves, demonstrating that the training converged stably without overfitting. 

These comparisons, however, assume full numerical precision, whereas any physical realization of the nonlinearities must operate at reduced bit-depth. Within \cref{fig:vit_results}\textcolor{blue}{d}, we report the sensitivity of test accuracy to decreasing quantization bit-depth. At 4-bit quantization, \textit{Optmax} achieves a mean test accuracy of \SI{74.6}{\percent}, compared to \SI{76.3}{\percent} for the Softmax benchmark. \textit{Optmoid} exhibits a more pronounced sensitivity to bit-depth, falling to \SI{69.9}{\percent} at 4 bits, whereas Sigmoid maintains \SI{75.9}{\percent}. We attribute this steeper degradation to the specific bias of $b = -4.16$ (see \cref{supp:simulations_details}), which, under restricted bit-depth, causes an excessive number of activations to be mapped to zero, effectively sparsifying the representation at the cost of relevant information. 

Additive noise, rather than quantization, emerged as the binding constraint on ViT accuracy. \cref{fig:vit_results}\textcolor{blue}{e} depicts the test accuracy for four ViTs (using \textit{Optmax} and \textit{Optmoid} with full precision and 4-bit quantization) under different noise levels. We note that here, we train the system without noise, and only during testing, we inject the noise on each \textit{Optmax}/\textit{Optmoid} element $s_i$ before the output quantization step:

\begin{equation}
    s_i = s_i + \mathcal{N}(0, \sigma^2)
\end{equation}

Our results show that under full precision, both \textit{Optmax} and \textit{Optmoid} maintain functionality even under strong noise conditions of $\sigma=0.1$, which was the maximum noise level we measured at $\SI{10}{GBaud}$ in \cref{fig:2}\textcolor{blue}{e}. However, after introducing 4-bit quantization, mean test accuracy already drops below $23\%$ at $\sigma=0.02$, effectively rendering the model non-functional. We further find that ViTs are much less susceptible to performance degradation under multiplicative noise than additive noise. We attribute this to attention outputs that were previously mapped to zero being lifted by additive noise above the lowest quantization threshold $1/2^4=0.0625$. The performance degradations under noise can further be mitigated by already accounting for noise during training. For a detailed analysis on training with noise and an additive-multiplicative comparison, we refer to \cref{supp:noise_modelling}.

\subsection*{Language Modeling}

\begin{figure*}[t]
    \centering
    \includegraphics[width=1\linewidth]{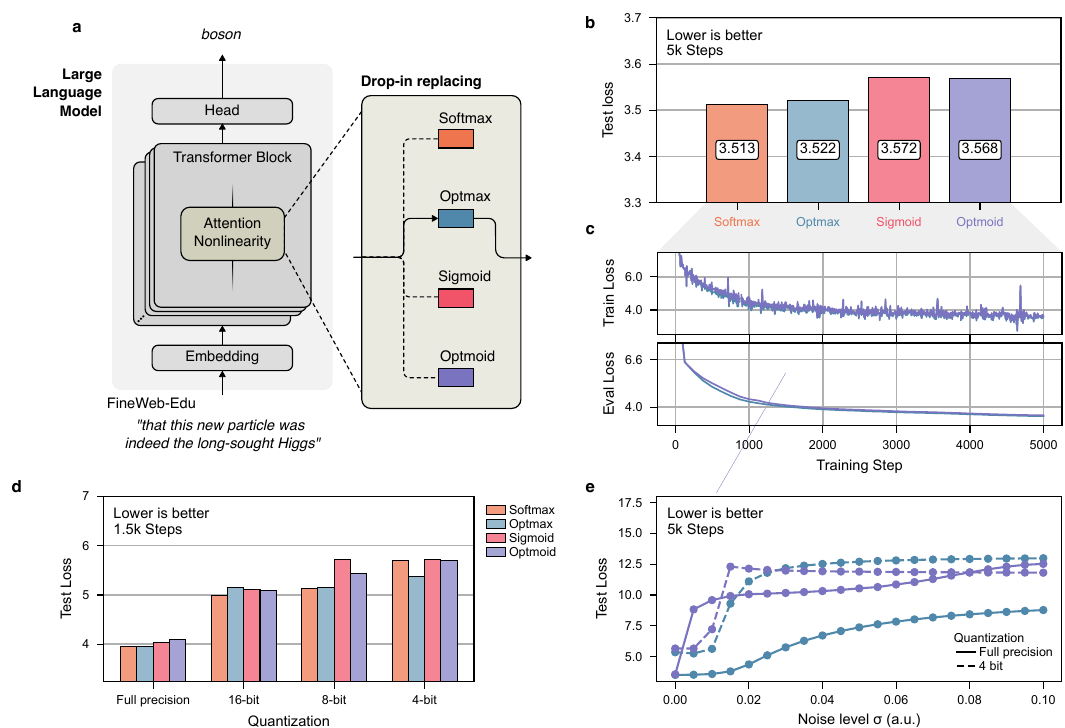}
    \caption{\textbf{GPT-2 FineWeb-Edu results using electro-optic nonlinearities}. \textbf{(a)} A large-language model predicts the next word-fragment in a sequence using stacked transformer blocks. We drop-in replace different attention nonlinearities.
    \textbf{(b)} Final test loss comparison for GPT-2 on FineWeb-Edu using 5k training steps. The proposed \textit{Optmax} and \textit{Optmoid} demonstrate competitive performance relative to standard Softmax and Sigmoid benchmarks. 
    \textbf{(c)} Training and validation loss curves for the \textit{Optmax} and \textit{Optmoid}, illustrating stable convergence. 
    \textbf{(d)} Quantization sensitivity analysis showing test loss as a function of bit-depth (full precision down to 4-bit) using 1.5k training steps.
    \textbf{(e)} Noise robustness showing final test loss under additive Gaussian noise $\mathcal{N}(0, \sigma^2)$ (no noise during training) using 2.5k training steps.}
    \label{fig:clm_results}
\end{figure*}

To test whether these results extend from vision to natural language tasks, we evaluated GPT-2 \citep{radford_language_2019} on FineWeb-Edu \citep{lozhkov_fineweb-edu_2024}. In this setting, the model processes sequences of tokens, which are typically short word fragments, through stacked transformer blocks to predict the next token in the sequence (\cref{fig:clm_results}\textcolor{blue}{a}). Given the preceding tokens $x_{1:t}$, the transformer, parameterized by $\theta$, models the conditional distribution of the next token $p_\theta(x_{t+1}|x_{1:t})$. To judge model quality, we evaluate the negative log-likelihood on unseen sequences of length $n$, which is given by $-\frac{1}{n-1}\sum_{i=1}^{n-1}\log p_\theta(x_{i+1} | x_{1:i})$ and quantifies how well the model predicts the correct next token. Here, lower loss values indicate better predictions. For a fair comparison, all variants share identical initializations and compute budgets, isolating the performance impact of the attention nonlinearities (see Methods).

Our results demonstrate that the proposed electro-optical nonlinearities remain highly competitive with standard digital alternatives, inducing only marginally higher test loss. As reported in \cref{fig:clm_results}\textcolor{blue}{b}, while Softmax obtains a test loss of $3.513$, \textit{Optmax} is competitive with a loss of $3.522$. Similarly, Sigmoid obtains a test loss of $3.572$, whereas \textit{Optmoid} achieves $3.568$. The corresponding validation and training loss curves are depicted in \cref{fig:clm_results}\textcolor{blue}{c}, demonstrating stable training across all variants.

\textit{Optmax} tracks Softmax closely across low-precision formats, even yielding a lower test loss at 4-bit quantization. Within \cref{fig:clm_results}\textcolor{blue}{d}, we analyze the sensitivity of the test loss after 1.5k training steps to decreasing quantization bit-depth. At $4$ bits, \textit{Optmax} yields a lower test loss than Softmax ($5.37$ versus $5.70$), while remaining essentially matched at $8$ bits ($5.15$ versus $5.12$). Similarly, \textit{Optmoid} proves as robust as Sigmoid under quantization, with a marginally lower loss at 4-bit ($5.71$ versus $5.72$) and a more pronounced advantage at $8$ bits ($5.44$ versus $5.72$). While Softmax remains slightly favorable in full-precision (bf16) relative to \textit{Optmax} and \textit{Optmoid}, the parity observed under quantization underscores the potential of these electro-optic nonlinearities for high-speed, low-bit accuracy hardware deployment. 

Language modeling proved markedly more noise-sensitive than image classification, degrading beyond $\sigma > 0.01$, below the measured noise floor at $\SI{10}{GBaud}$. In \cref{fig:clm_results}\textcolor{blue}{e}, we again report \textit{Optmax} and \textit{Optmoid} performance with additional additive noise in inference in full precision and 4-bit quantization. This functional threshold is substantially lower than the additive noise observed in experiment at $\SI{10}{GBaud}$ ($\sigma=0.098$ for \textit{Optmax} and $\sigma=0.088$ for \textit{Optmoid}). Under full precision, \textit{Optmax} test loss remains below $5$ for $\sigma<0.02$, while \textit{Optmoid} already breaks the functional threshold for $\sigma=0.01$. Under $4$ bit, both remain functional for $\sigma<0.01$, leveling out to losses of around $12.5$ afterwards. We emphasize that this does not represent a functional model at $\sigma=0.1$; rather, such high loss values are indicative of a poorly performing or untrained GPT-2.

Together, the ViT and GPT-2 benchmarks show that bounded nonlinearities like the MZM sinusoidal are not a barrier to transformer-level performance, isolating additive noise as the main outstanding obstacle.

%% file: 04_discussion.tex
\label{discussion}
\begin{table*}[t]
    \centering
    \caption{Performance Comparison: Latency and Energy consumption for a sequence length $n=64$. \textit{Optmax} and \textit{Optmoid} values are calculated at $\SI{10}{GBaud}$.}
    \label{tab:performance_comparison}
    \begin{tabular}{l l l l l l}
        \toprule
        \textbf{Architecture} & \makecell{\textbf{Latency (s)} \\ \textbf{per Sequence}} & \makecell{\textbf{Energy (J)} \\ \textbf{per Sequence}} & \textbf{Type} & \textbf{Technology} & \textbf{Ref} \\
        \midrule
        nMOS SMA & 5.5e-04 & 1.9e-08 & Electronic & Analog & \cite{sillman2023analogimplementationsoftmaxfunction} \\
        Softermax & 7.7e-04 & 1.3e-08 & Electronic & Digital & \cite{Stevens_Softermax} \\
        Softonic & 1.7e-05 & \textbf{4.5e-11} & Optic & Analog & \cite{dash_softonic_2025} \\
        VEXP & 2.2e-07 & 5.0e-08 & Electronic & Digital & \cite{wang_vexp_2025} \\
        \addlinespace[0.5em]
        \textit{Optmax} & 1.3e-08 & 1.0e-08 & Electro-Optic & Analog & This work \\
        \textit{Optmoid} & \textbf{6.5e-09} & 4.7e-09 & Electro-Optic & Analog & This work \\
        \bottomrule
    \end{tabular}
\end{table*}

The central contribution of this work is the repurposing of nonlinear optic modulators as accelerators for nonlinear attention functions at the heart of transformers. We show that a formerly digital primitive, Softmax, can be replaced by the analog response of standard telecom modulators with highly competitive model accuracy.

The switch to high-throughput analog optics, however, introduces a constraint that is true of all analog computation: an analog system's bounded output range does not permit an exact representation of the unbounded exponential and reciprocal functions in Softmax. In the case of the MZM, transmission is inherently periodic and capped at unity, with the system's noise floor further setting the lower bound. Overall, these effects lead to a compressed output range and \textit{Optmax}'s departure from the strict unit-sum constraint of Softmax.

Despite these limitations, approximations remain highly effective in both vision and language tasks. We attribute this to the underlying mechanisms of the attention operation itself. The empirical success of Softmax attention relies primarily on two key properties: the non-negativity of the attention matrix and a non-linear re-weighting scheme that concentrates its distribution \cite{zhen2022cosformer}. Despite not capturing the full dynamic range of exponential terms, our MZM-based approximations therefore do preserve these two essential properties.

A significant finding is that in GPT-2, \textit{Optmax} and \textit{Optmoid} retain model functionality even under strong 4-bit quantization. In digital regimes, Softmax is typically the component most susceptible to accuracy loss during quantization \cite{kim2021bert} due to the sensitivity of the summation process \cite{pandey_softmax_2023}. Our architecture sidesteps this by isolating quantization to the input (DAC) and output (ADC) stages. While the data interface is low-precision, the internal summation occurs entirely in the analog domain. Within this analog backbone, computation proceeds with theoretically arbitrary physical precision (limited by noise rather than bit-width), bypassing the rounding errors that plague digital 4-bit fixed-point math. This gives a clear avenue for high-speed systems at which DAC/ADC stages readily available in low-precision formats. While our evaluation relies on models trained from scratch, future work must assess whether these performance values translate to the fine-tuning of larger pretrained transformers.

Our analysis of physical noise robustness highlights that additive noise remains the primary driver of performance degradation. When the noise floor interacts with the quantization threshold, it inadvertently activates suppressed weights, corrupting the attention distribution. In contrast, we report that the system is substantially more resilient when training with noise or when the noise source is purely multiplicative (see \cref{supp:noise_modelling}). These results suggest an inherent robustness to the multiplicative effects, like gain fluctuations, typical of photonic circuits. Going forward, hardware efforts should therefore focus on minimizing additive noise, while incorporating noise-aware training regimes could further bridge the gap between simulation and practical analog deployment.

Accounting for the full signal chain, we project \textit{Optmax} and \textit{Optmoid} to reduce latency by more than one order of magnitude compared to reported accelerators (\cref{tab:performance_comparison}). This includes DACs, ADCs, RF-amplifiers, thermal-biasing, and photodetectors rated at $\SI{10}{GBaud}$ (see \cref{supp:latency,supp:power}).  While some micro-ring resonator designs, such as SOFTONIC \cite{dash_softonic_2025}, report lower energy consumption per sequence, they face significant scaling challenges for larger sequence lengths. Our approach prioritizes a drastic reduction in latency while maintaining competitive power efficiency, recognizing that Softmax energy costs are often negligible compared to the total system compute power. To meet this goal, fully optical DACs \cite{song_integrated_2025} and ADCs \cite{11358960} could further mitigate overall latency by bypassing bandwidth-limited electrical conversion stages.

With these findings, we aim to stimulate exploration of bounded physical nonlinearities as a general substrate for attention. Other co-packaged optical technologies could similarly be identified to provide comparable functionality. Indeed, our approach is not tied to the Mach–Zehnder modulator: any steep, bounded response — a micro-ring's Lorentzian edge, an electro-absorption modulator's exponential roll-off, or vertical-cavity surface-emitting laser drive dynamics — could serve the same role and merit comparative studies. It remains to be determined whether periodically-poled TFLN waveguides operated in pump depletion could even extend Sigmoid-like attention to small-scale all-optical transformers. For electro-optic devices like the ones studied here, closing the simulation-to-deployment gap will require characterizing additive noise within a fully co-packaged signal chain rather than in isolated devices.

\newpage

%% file: 05_methods.tex
\label{Methods}

%During the preparation of this manuscript, the authors used commercial LLMs to revise original writing for to clarity and readability. The illustrations of a car, waveform and a robot in \cref{fig:attention-figure}\textcolor{blue}{a} were created using Google's \textit{Nano Banana}.

\subsection*{Device fabrication}
The measured thin-film lithium niobate MZMs are fabricated on a commercially available \SI{300}{\nm} thick, $5 \%$ magnesium oxide-doped lithium niobate thin film on a \SI{2}{\um} thick silicon dioxide insulation-layer on a silicon handle.
The optical layer is defined by electron beam lithography using hydrogen silsesquioxane resist and etched into the thin-film using an optimised argon etching process \cite{FabPaper, sabatti_extremely_2024}. A cleaning step is then performed with KOH to remove the etching by-products and with buffered HF to remove the remaining mask. The RF electrodes are defined by direct laser writing lithography, a standard lift-off process and electron beam evaporation of \SI{850}{\nm} of Au with a \SI{5}{nm} thick Cr adhesion layer.

\subsection*{High-speed RF characterization}
 
We characterize the electro-optic response of a fabricated MZM against the differentiable model used for transformer training. A 5-bit uniformly sampled sequence of length $n=2048$ was loaded into a Micram DAC10002 arbitrary-waveform generator (\SI{100}{GS/s}); the single-ended RF output was AC-coupled through an SHF BT45R \SI{45}{GHz} bias-tee and amplified by an AT-Microwave AT-LNA-0043-3504Y (\SI{35}{dB}, \SI{43.5}{GHz}). A Toptica CTL 1500 laser (\SI{1550}{nm}, \SI{30}{mW} fiber output) provided the optical carrier, and an erbium-doped fiber amplifier (Keopsys CEFA-C-PB-HP) compensated fiber-to-chip coupling loss at the MZM output. The modulated light was detected by a Thorlabs DXM50AF photodiode and recorded on a Tektronix DPO77002SX (\SI{200}{GS/s}) real-time oscilloscope. For \textit{Optmoid}, the drive was set to $V_{pp}=V_\pi=\SI{5.73}{V}$ with the thermal phase shifter biased at quadrature; for \textit{Optmax}, $V_{pp}=V_\pi/2=\SI{2.87}{V}$ with the bias set midway between the minimum-transmission and quadrature points. Sequences were encoded at \SI{100}{MBaud}, \SI{1}{GBaud}, and \SI{10}{GBaud} (DAC operated at \SI{80}{GS/s} with 8 and 80 samples per symbol for \SI{10}{} and \SI{1}{GBaud}, and at \SI{8}{GS/s} with 80 samples per symbol for \SI{100}{MBaud}). Captured traces were low-pass filtered (\textit{scipy.signal.firwin} \cite{noauthor_firwin_nodate}, cut-offs \SI{12}{GHz}, \SI{1.2}{GHz}, \SI{120}{MHz}), decimated to 20 samples per symbol, and integrated over a window of $20\%$ of the symbol period to emulate a triggered receiver. A setup sketch, full waveforms, and a noise-source breakdown are given in \cref{Supp:ghz_noise_analysis}.
 
\subsection*{Model training, quantization, and noise emulation}
 
Both nonlinearities were implemented as differentiable JAX- and PyTorch-compatible modules and inserted as drop-in replacements for Softmax and Sigmoid, leaving all other components of the transformer models unchanged. Finite DAC/ADC resolution was emulated by quantizing the nonlinearity input and output, and the identical quantization was applied to the output of Softmax and Sigmoid baselines for a fair comparison. Measured analog fluctuations were reproduced by injecting Gaussian noise before output quantization, either multiplicatively or additively. A further analysis of the impact of noise on the attention-mechanism is given in \cref{supp:noise_modelling}.
 
For image classification, we trained a ViT \cite{dosovitskiy_vit_2021} (256 embedding dimension, 512 hidden dimension, 8 heads, 6 layers, patch size 4, dropout 0.2) at fp32 precision on MNIST, CIFAR-10, and SVHN \cite{mnist_2010, cifar_2009, svhn}, swapping only the attention nonlinearity. Hyperparameters were tuned on the CIFAR-10 validation split (200 epochs, batch size 128, learning rate $5\times10^{-4}$); final accuracies were reported on a held-out test set averaged over three seeds. 

For causal language modeling we trained GPT-2 (124M) \cite{radford_language_2019} from scratch on FineWeb-Edu \cite{lozhkov_fineweb-edu_2024} with a 1024-token context, AdamW (weight decay 0.1, learning rate $6\times10^{-4}$, attention dropout $0.1$), and a warmup-stable-decay schedule ($5\%$ warmup, $28.5\%$ decay) in bf16 mixed precision, with an effective batch of 512 sequences ($\approx$\SI{0.5}{M} tokens per step). The main comparison used 5{,}000 steps ($\approx$\SI{2.6}{B} tokens); the quantization ablation used 1{,}500 steps with symmetric 4-, 8-, and 16-bit input/output quantization in \textit{Optmax}/\textit{Optmoid} and output quantization for Softmax/Sigmoid. The noise study used 2{,}500 steps. Data was split 94/5/1 into train/validation/test; validation was performed every 125 steps on 50 batches and final test metrics on 200 batches. Model quality was quantified by the token-level negative log-likelihood defined in the Results. Further configuration details and hyperparameter sweeps are tabulated in \cref{supp:simulations_details}.

\subsection*{\textit{Optmax} Calibration}
\label{Supp_calibration}
%\subsection{\textit{Optmax} Calibration}

\textit{Optmax} approximates the standard Softmax function, $\text{Softmax}(\mathbf{x})_j = e^{x_j} / \sum_{i=1}^{n} e^{x_i}$, by decomposing the operation into an analog exponentiation stage and a subsequent normalization stage:
\begin{equation}
    \text{Optmax}(\mathbf{x})_i \approx  f_{\exp}(x_i) \cdot N(z),
\end{equation}
where $f_{\exp}$ denotes the exponential approximation and $N(z)$ represents the normalization factor applied to the accumulated sum $z = \sum_j f_{\exp}(x_j)$. The fidelity of approximation is governed by two primary hyperparameter domains that define the training simulations:
\begin{enumerate}
    \item The input encoding range $[x_{\min}, x_{\max}]$, which maps onto the rising slope of the first MZM's transfer function.
    \item The normalization range $[z_{\min}, z_{\max}]$, which maps the accumulated optical power onto the falling slope of the second MZM.
\end{enumerate}

To translate from the digital simulation domain $x \in [x_{\min}, x_{\max}]$ to the physical voltage input $V \in [V_{\min}, V_{\max}]$ driving the MZM, we define a fixed, range-preserving affine transformation:
\begin{equation}
    V(x) = \gamma x + \delta,
\end{equation}
where the scaling factor $\gamma$ and the bias $\delta$ are given by:
\begin{align}
        \gamma &= \frac{V_{\max} - V_{\min}}{x_{\max} - x_{\min}},\\
    \delta &= V_{\min} - \gamma \cdot x_{\min}.
\end{align}

We determine the physical transmission function $T(V)$ of the MZM via a fit of the measured experimental data over the voltage window $V \in [V_{\min}, V_{\max}]$, modeled as:
\begin{equation}
    T(V) = a(1+\sin(bV + c)).
\end{equation}

To approximate the exponential numerator of the Softmax, the digital input $x$ is mapped strictly to the rising slope of the MZM's sinusoidal transfer function. The physical voltage boundaries are selected such that $V_{\min}^{(\mathrm{exp})}$ corresponds to the point of minimum transmission and $V_{\max}^{(\mathrm{exp})}$ corresponds to the point of maximum positive gradient in the measured data. Fit and resulting approximation $f_{\exp}(x)$ for $[x_{\min}, x_{\max}] = [0, 4]$ are shown in Supplementary \cref{fig:S1}\textcolor{blue}{a--b}.

The normalization factor $N(z)$ accounts for the falling slope of the second MZM, a global proportionality factor $\alpha$ of the output ADC, and the finite extinction ratio $\beta \in [0, 1]$ of the second MZM:
\begin{equation}
    N(z) = \alpha \cdot \left[ \beta + (1-\beta) f_{\text{rec}}(z) \right].
    \label{eq:s_fit_function}
\end{equation}
Here, $f_{\text{rec}}$ represents the normalized transmission along the falling slope within the defined domain $z \in [z_{\min}, z_{\max}]$. The voltage boundaries are selected such that $V_{\min}^{(\mathrm{rec})}$ aligns with the maximum negative gradient and $V_{\max}^{(\mathrm{rec})}$ aligns with the minimum transmission point. We optimize $N(z)$ in $\alpha$ and $\beta$ to minimize the discrepancy between the physical response and the ideal reciprocal function $1/z$. Supplementary \cref{fig:S1}\textcolor{blue}{d} exemplifies the calibrated $N(z)$ for the range $[z_{\min}, z_{\max}] = [6, 14]$.

We model the finite bit-depth of the DAC and ADC using two quantization functions: $q_{\text{in}}(\cdot)$ for the DAC stage and $q_{\text{out}}(\cdot)$ for the ADC stage. 

Furthermore, we account for the stochastic nature of the analog system by introducing either an additive or multiplicative Gaussian noise term before the output quantization step. For multiplicative noise, the complete differentiable forward model for \textit{Optmax} used during training is:

\begin{equation}
    \begin{aligned}
        \text{Optmax}(x)_i = q_{\text{out}} \Big( & N(z) \cdot f_{\exp}\left(q_{\text{in}}(x_i)\right) \times\mathcal{N}(1, \sigma^2) \Big) \;,
    \end{aligned}
\end{equation}

and for additive noise similarly, 

\begin{equation}
    \begin{aligned}
        \text{Optmax}(x)_i = q_{\text{out}} \Big( & N(z) \cdot f_{\exp}\left(q_{\text{in}}(x_i)\right) + \mathcal{N}(0, \sigma^2)\Big) \;.
    \end{aligned}
\end{equation}
During training, we either set $\sigma=0$ and evaluate noise robustness only during inference, or we also train with $\sigma>0$.

\subsection*{\textit{Optmoid} Calibration}

The \textit{Optmoid} unit approximates the element-wise Sigmoid function $\text{Sigmoid}(\mathbf{x})_j = 1/(1+e^{-(x_j+b)})$ by exploiting the full min-to-max swing of the MZM sinusoidal response:
\begin{equation}
    \text{Sigmoid}(x_j + b) \approx f_{\text{sig}}(x_j + b),
\end{equation}
where $b$ is a sequence-length-dependent bias hyperparameter. The voltage boundaries of the mapping are selected such that $V_{\min}^{(\mathrm{exp})}$ and $V_{\max}^{(\mathrm{exp})}$ correspond to the points of minimum and maximum optical transmission, respectively. We calibrate the input scaling within the bounds $[x_{\min}, x_{\max}]$ to optimally match the ideal Sigmoid distribution for a given bias via a least-squares fit. Supplementary \cref{fig:S1}\textcolor{blue}{e} depicts the resulting transfer function $f_{\text{sig}}(x)$ with $b=-3.93$.

To simulate the physical constraints of the integrated system, we again model the finite bit-depth of the DAC and ADC interfaces using two quantization functions, and account for experimental noise by introducing a Gaussian noise term before the final digitization stage. 

The complete differentiable forward model for multiplicative noise is given by:
\begin{equation}
    \begin{aligned}
        \text{Optmoid}(x)_i = q_{\text{out}} \Big( & f_{\text{sig}}\left(q_{\text{in}}(x_i)\right) \times\mathcal{N}(1, \sigma^2) \Big) \;,
    \end{aligned}
\end{equation}

and for additive noise, respectively

\begin{equation}
    \begin{aligned}
        \text{Optmoid}(x)_i = q_{\text{out}} \Big( & f_{\text{sig}}\left(q_{\text{in}}(x_i)\right) + \mathcal{N}(0, \sigma^2)\Big) .
    \end{aligned}
\end{equation}

We again either set $\sigma>0$ only during evaluation or during training and evaluation.

%% file: xx_supplementary.tex
\setcounter{figure}{0}
\renewcommand{\thefigure}{S\arabic{figure}}

\section{Softmax Bottleneck on GPUs}
\label{supp:softmax_bottleneck}

\begin{figure}[t]
    \centering
    \includegraphics[width=\linewidth]{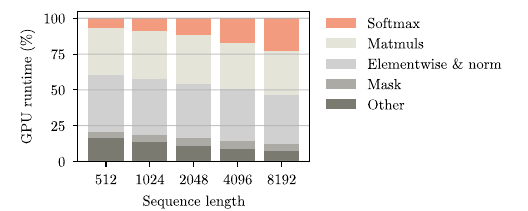}
    \caption{GPU runtime of GPT-2 for different sequence lengths broken down by operation. To enable a clear attribution of runtime within scaled dot product attention, we use the PyTorch MATH backend \cite{noauthor_sdpbackend_nodate}.}
    \label{fig:s-1_runtime}
\end{figure}

Computing one forward pass of an LLM involves a combination of linear and nonlinear operations. The total compute budget is typically measured in floating-point operations (FLOPs) and heavily dominated by linear operations such as general matrix multiplications (GEMMs). At sequence length $n=8192$, GEMMs account for \SI{99.43}{\%} of total FLOPs within the GPT-2 model, while the Softmax operations account for only \SI{0.56}{\percent} \cite{radford_language_2019, hoffmann2022training, kaplan2020scalinglawsneurallanguage}. To obtain an estimate for the runtime impact of the Softmax, we measure the execution time of a GPT-2 forward pass on an NVIDIA H100 GPU and break it down by operation. We present the results in \cref{fig:s-1_runtime} where the Softmax alone accounts for \SI{22}{\percent} of total execution time for a sequence length of $n=8192$. We remark that here, to attain a clean runtime breakdown of the transformer's components, we leverage scaled dot-product attention (\cref{fig:s-1_runtime}) using the PyTorch MATH backend \cite{noauthor_sdpbackend_nodate, paszke_pytorch_nodate}. In practice, the individual operations within attention are often fused into a single GPU function (kernel), which can alleviate this bottleneck significantly by avoiding repeated read and write operations to memory \cite{dao2022flashattention}. Nevertheless, even with this optimization, the problem of a comparatively slow Softmax remains, as the throughput for the exponential function on modern GPUs is significantly lower than that of linear multiply-accumulate operations. For an NVIDIA H100 GPU, the exponential function (exp2) throughput is about $256\times$ lower than that of linear matrix multiply-accumulate (MMA): each streaming multiprocessor in an H100 supports 4096 MMA operations but only 16 exp2 operations per clock cycle. For the most recent NVIDIA GPUs, the GB200 and GB300, this throughput disparity even increases to $512\times$ and remains at $256\times$, respectively \cite{zadouri2026flashattention4algorithmkernelpipelining}.

At the time of writing, state-of-the-art software acceleration techniques go so far as to approximate exponential functions using piecewise polynomial expansions \cite{zadouri2026flashattention4algorithmkernelpipelining}, enabling their execution on faster linear compute units. When assuming a key/query projection dimension of $d_k=128$ per attention head, the aforementioned throughput disproportionality implies that the time required to evaluate the exponential function is roughly $50\%$ of that of the matrix multiplications within the attention block \cite{shah2024flashattention}. Through the adoption of lower precision numerical formats, this percentage is only going to increase further \cite{wang_sole_2023}. 

\begin{figure*}[t]
    \centering
    \includegraphics[width=1\linewidth]{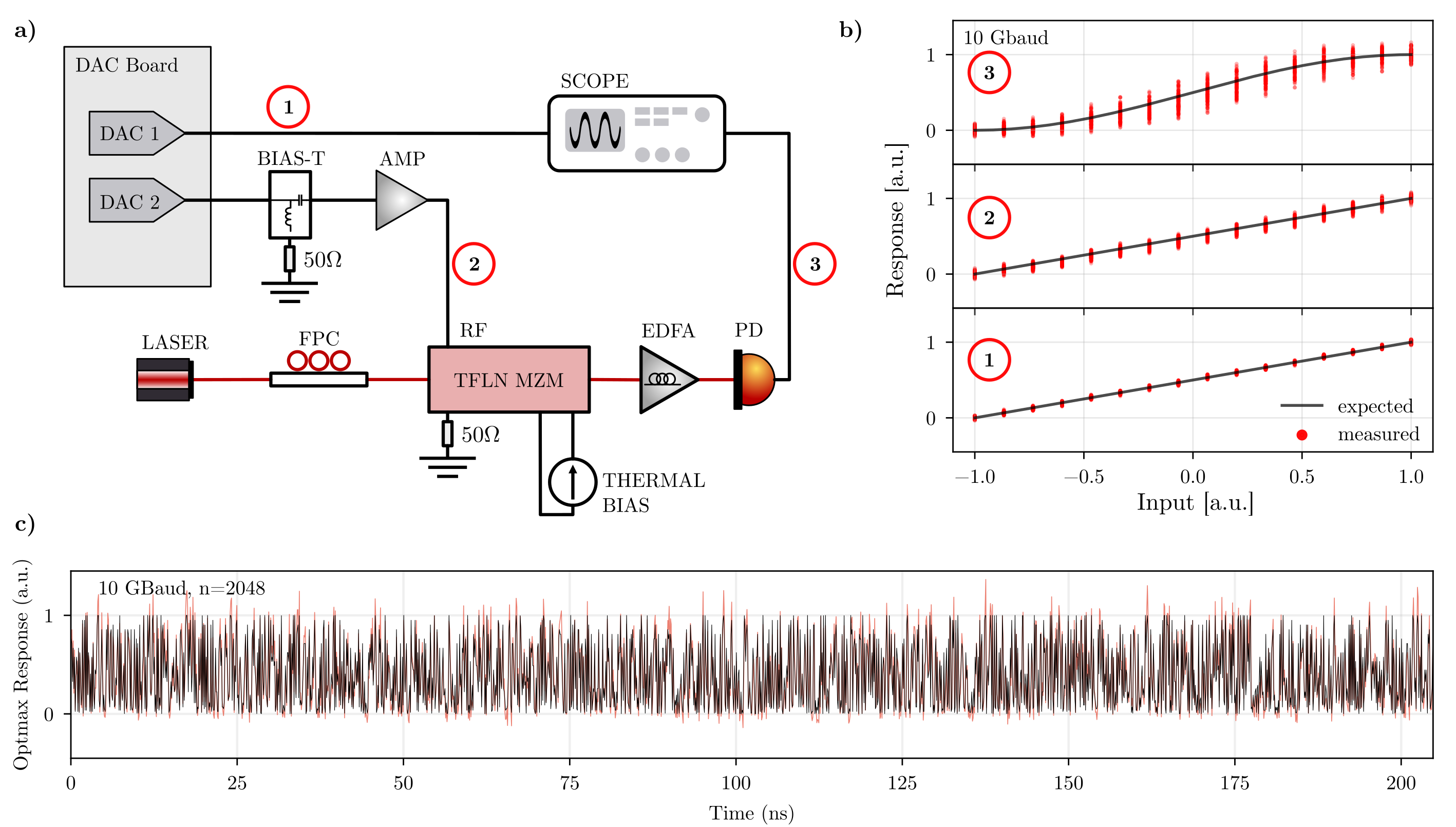}
    \caption{Experimental setup and signal waveforms.
\textbf{(a)} Schematic of the MZM characterization setup, including the RF driving chain and optical detection path. 
\textbf{(b)} Time-domain traces of a \SI{10}{GBaud} sequence at the DAC output, RF amplifier output, and photodiode output. Noise is maximized at the MZM quadrature point where voltage-to-power transconductance is highest. \textbf{(c)} A full time trace of a $n=2048$ \textit{Optmax} response under \SI{10}{GBaud}.}
    \label{fig:s0_setup}
\end{figure*}

\section{Setup and Measurement}
\label{Supp:ghz_noise_analysis}

We perform high-speed measurements of a Mach-Zehnder modulator (MZM) response to compare the experiment with the simulations used within transformer training. \cref{fig:s0_setup}\textcolor{blue}{a} illustrates the experimental setup described in Methods. 

\cref{fig:s0_setup}\textcolor{blue}{b} displays the response of a uniformly sampled \SI{10}{GBaud} 4-bit sequence at $V_\pi$ at three stages: (1) DAC output, (2) RF amplifier output, and (3) photodiode output. We observe additive noise introduced by the RF amplifier, which is most pronounced at the MZM quadrature point where the voltage-to-power transconductance is maximized. Following photodetection, the noise level increases further, indicating combined contributions from amplified spontaneous-emission beat noise originating from the erbium-doped fiber amplifier (EDFA), as well as the thermal noise floor of the photodiode. Further research is required to distinctly characterize the noise sources.

\cref{fig:s0_setup}\textcolor{blue}{c} displays a full time series of the normalized \textit{Optmax} response for a randomly sampled input sequence of length $n=2048$ at \SI{5}{bit} resolution. The figure draws a comparison between the simulation used in our transformer training and experimental measurements. For the latter, we measure a thin-film lithium niobate MZM's rising slope driven at \SI{10}{GBaud}. The greyed-out section indicates the zoomed-in segment of the main manuscript \cref{fig:2}\textcolor{blue}{c}.

\begin{figure*}
    \centering
    \includegraphics[width=1\linewidth]{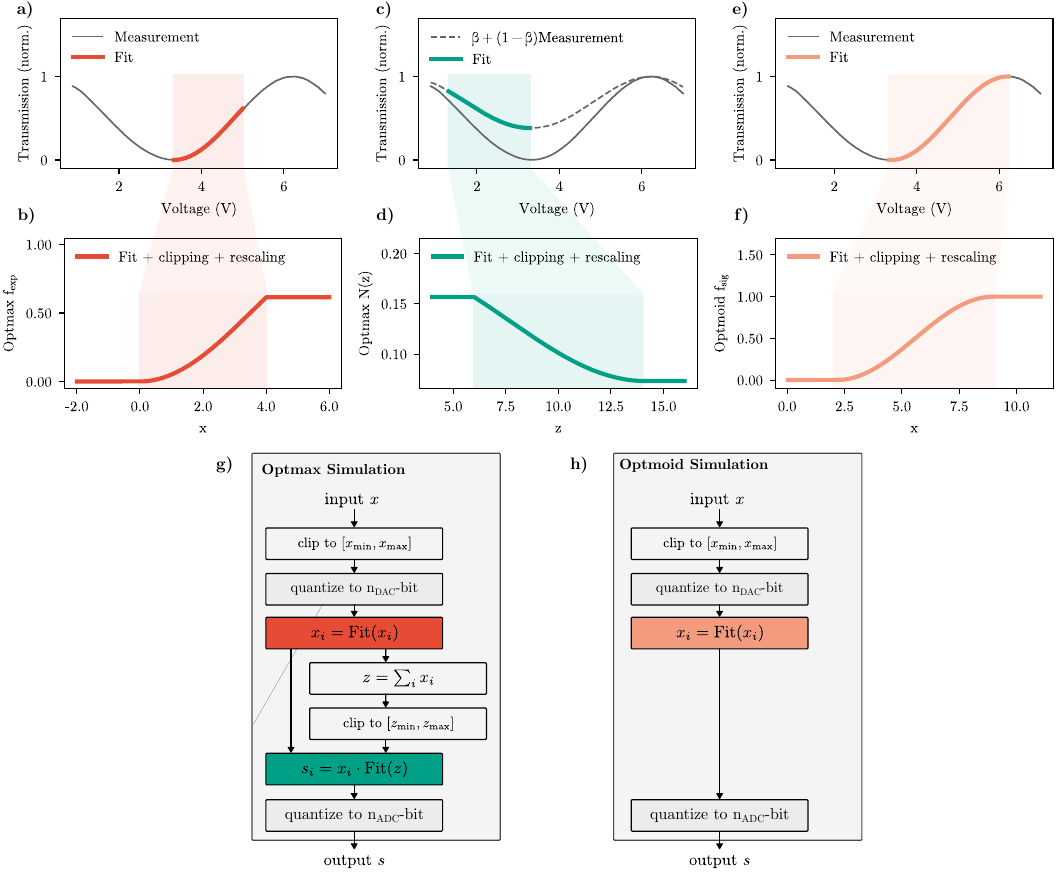}
    \caption{Calibration of \textit{Optmax} and \textit{Optmoid} electro-optic units.
    \textbf{(a--b)} Least-squares fit of MZM transmission $T(V)$ and the resulting exponential approximation $f_{\exp}(x)$ mapped to the rising slope for $x \in [0, 4]$. 
    \textbf{(c--d)} Characterization of the normalization stage MZM and the optimized factor $N(z)$ for $z \in [6, 14]$. The fit minimizes error against $1/z$ by tuning gain $\alpha$ and extinction ratio $\beta$. 
    \textbf{(e)} \textit{Optmoid} transfer function $f_{\text{sig}}(x)$ using the full min-to-max MZM swing with bias $b=-3.93$. 
    Solid lines represent hardware-aware forward models (including quantization $q_{\text{in}}, q_{\text{out}}$), while dashed lines indicate ideal mathematical targets. \textbf{(g-h)} Simulation frameworks of \textit{Optmax} and \textit{Optmoid} without noise injection.}
    \label{fig:S1}
\end{figure*}

\section{Calibration Details}
\label{supp:simulation}

The calibration of the \textit{Optmax} and \textit{Optmoid}, the MZM transmission fits and the resulting $f_{\exp}$, $N(z)$, and $f_{\text{sig}}$ approximations described in Methods are shown in \cref{fig:S1}\textcolor{blue}{a-f}. Fits were performed with 

$$
T(V) = a(1+\sin(bV + c)).
$$
For the rising, lowering, and full-swing, the fit parameters used in training are:
\begin{equation}
\begin{aligned}
(a,b,c,V_{\min},V_{\max}) &= (0.523,\,1.04,\,1.26,\,3.333,\,5.000),\\
(a,b,c,V_{\min},V_{\max}) &= (0.507,\,1.01,\,1.40,\,1.316,\,3.285),\\
(a,b,c,V_{\min},V_{\max}) &= (0.500,\,1.11,\,0.959,\,3.333,\,6.228).
\end{aligned}
\end{equation}

Depicted in \cref{fig:S1}\textcolor{blue}{g}, \textit{Optmax} first clips input $x_i$ to $[x_{\min}, x_{\max}]$ after which each value is quantized to the desired bit-depth - emulating the behavior of an input digital-to-analog converter (DAC). The first fit then simulates a modulation on top of the rising swing of an MZM. The full sequence is then summed and again clipped to $[z_{\min}, z_{\max}]$, after which the second fit-function simulates the behavior of an MZM modulated at the lowering swing. The output is again quantized and returned.

The \textit{Optmoid} simulation, depicted in \cref{fig:S1}\textcolor{blue}{h}, simply skips the summing, second clipping and fit step.

\section{Noise in Training}
\label{supp:noise_modelling}
\begin{figure*}
    \centering
    \includegraphics[width=1\linewidth]{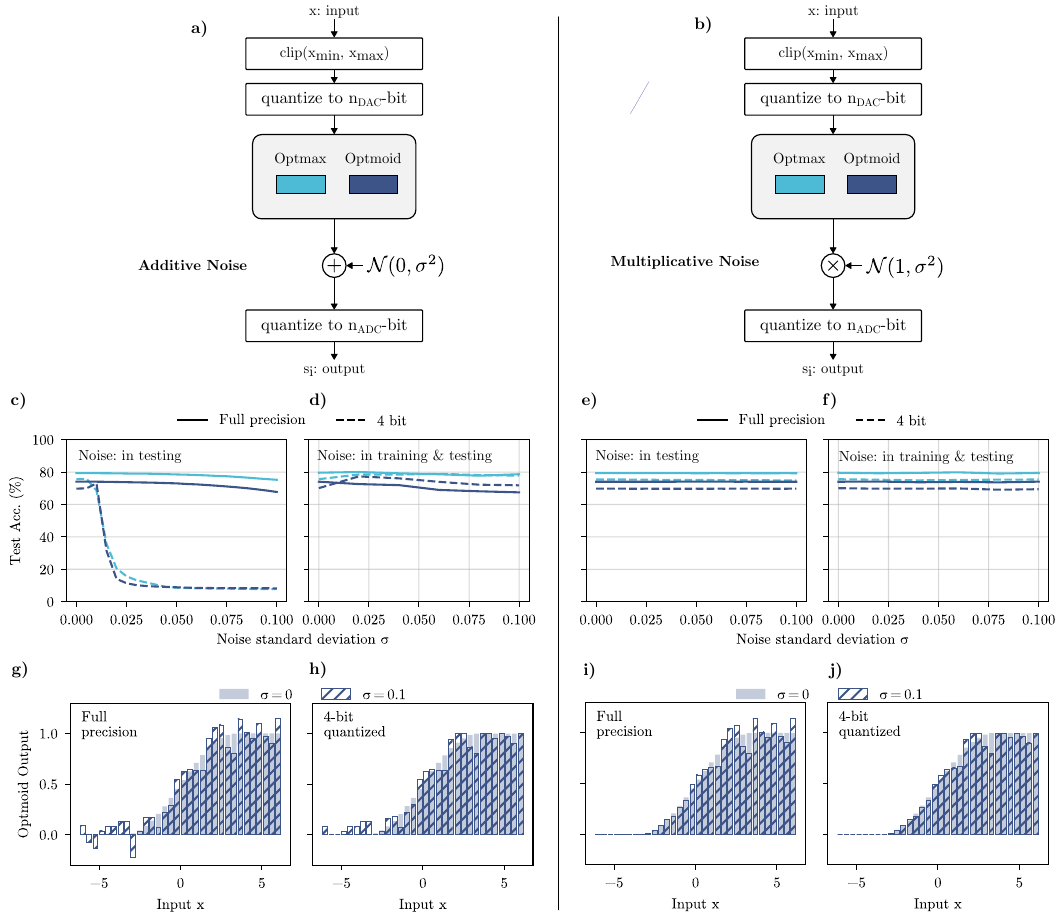}
    \caption{\textbf{Noise effects in \textit{Optmax} and \textit{Optmoid} for ViT training} \textbf{(a)} Additive noise model. \textbf{(b)} Multiplicative noise model. \textbf{(c)} Test accuracy versus additive noise applied only during inference, showing strong degradation under 4-bit quantization. \textbf{(d)} Test accuracy with additive noise during both training and inference, demonstrating improved robustness and performance. \textbf{(e)} Test accuracy under multiplicative noise applied only during inference, showing minor degradation. \textbf{(f)} Test accuracy under multiplicative noise during both training and inference, indicating stable performance. \textbf{(g)} Additive noise in \textit{Optmoid} without output quantization, producing out-of-range values. \textbf{(h)} Additive noise in \textit{Optmoid} with output quantization, illustrating threshold lifting effects. \textbf{(i)} Multiplicative noise in \textit{Optmoid} in full precision, preserving non-negativity. \textbf{(j)} Multiplicative noise in \textit{Optmoid} with quantization, maintaining zeroed values.}
    \label{fig:S2}
\end{figure*}

\cref{fig:S2}\textcolor{blue}{a} and \textcolor{blue}{b} illustrate the difference between additive and multiplicative noise when using \textit{Optmax} or \textit{Optmoid} simulations in transformer training. When training the Vision Transformer (ViT) as reported in Results, we observe noticeable differences whether we account for additive or multiplicative noise. 

In \cref{fig:S2}\textcolor{blue}{c} we show the impact of additive noise on the final test accuracy under full-precision or 4-bit quantization. The plot reports noise values from $\sigma=0$ to $\sigma = 0.1$, but accounts for noise only in testing, not in training. Under full-precision, the test-accuracy degrades from $75.34\%$ to $75.23\%$ for \textit{Optmax} and from $74.06\%$ to $67.72\%$ for \textit{Optmoid}. Under 4-bit quantization, however, accuracies sharply degrade to as low as $21.28\%$ for \textit{Optmax} and $14.51\%$ for \textit{Optmoid} at a noise level of only $\sigma=0.02$. We note that in our high-speed $\SI{10}{GBaud}$ experiments in Results we measure much higher additive noise at values of $\sigma=0.098$ and $\sigma=0.088$ for \textit{Optmax} and \textit{Optmoid}, respectively. These results indicate that training without noise and under 4-bit quantization is not robust for noise levels beyond $\sigma>0.02$. 

\cref{fig:S2}\textcolor{blue}{d} shows the test accuracy under additive noise, but in this case also accounting for noise whilst training. We report a much more stable scenario, indicating that training with noise could be highly beneficial. Under 4-bit quantization and at $\sigma = 0.1$, \textit{Optmax} even improves from $75.59\%$ to $77.77\%$ and \textit{Optmoid} from $70.07\%$ to $71.81\%$. Investigating how and why this noisy and quantized attention mechanism improves in this scenario will necessitate further research. Within this work, similar noise-in-training analyses were not conducted with GPT-2 due to limitations in compute availability. 

\cref{fig:S2}\textcolor{blue}{g} and \textcolor{blue}{h} illustrate additive noise at the example of \textit{Optmoid}. Panel \textcolor{blue}{g} shows the output under additive noise $\sigma=0.1$ and no noise with full precision. In this setting, \textit{Optmoid} can output values $s_i<0$ and $s_i>0$, which is not the designed intention of a strictly positive attention nonlinearity. However, this does not relate to a realistic physical scenario: input and output quantization (DAC and ADC) cannot be circumvented in a hybrid digital-analog setup. \cref{fig:S2}\textcolor{blue}{h} shows the output during the realistic case (4-bit quantization) using exactly the same noise key. Now, \textit{Optmoid} values are again within the desired range $[0,1]$. Crucially, some outputs for low $x$ are still non-zero. In the no-noise ($\sigma=0$) scenario the same $x$ values would be mapped to zero (shown in a light blue backdrop). We attribute most of the measured model degradation to this phenomenon: attention values which, under no noise, would be mapped to zero can suddenly, under noise, participate in the overall attention output. This phenomenon is not present during purely multiplicative noise, which is elaborated in the following.

\cref{fig:S2}\textcolor{blue}{e} and \textcolor{blue}{f} show the final test accuracy when testing or training with multiplicative noise. In both scenarios, the model maintains robust functionality. Under 4-bit quantization \textit{Optmax} degrades from $75.42\%$ to $74.94\%$ and \textit{Optmoid} $69.77\%$ to $69.82\%$ when accounting for noise in testing. For noise in training and testing, \textit{Optmax} degrades from $75.61\%$ to $75.44\%$ and \textit{Optmoid} $70.07\%$ to $69.44\%$. These results indicate that also in the multiplicative scenario, training with noise is beneficial. It does not, however, bear the same effect as under additive noise.

\cref{fig:S2}\textcolor{blue}{i} and \textcolor{blue}{j} illustrate multiplicative noise. Panel \textcolor{blue}{i} shows the output under multiplicative noise $\sigma=0.1$ and no noise with full precision. In this setting \textit{Optmoid} does not output values $s_i<0$, which is in contrast to the additive case (panel \textcolor{blue}{g}). The same is the case for the 4-bit quantized setup, shown in panel \textcolor{blue}{j}. More importantly, $x$ values which are mapped to zero under no noise (light blue backdrop) are also mapped to zero under multiplicative noise. 

\section{Latency}
\label{supp:latency}

We define the latency as the time required to compute the nonlinear operation over a sequence of length \(n\), where \(n\) represents the sequence length corresponding to the context size of the transformer.

The system consists of a cascade of stages: DAC conversion, optical modulation and propagation, photodetection (PD) with transimpedance amplification (TIA), and ADC conversion.

Let $T_{\mathrm{DAC}}$, $T_{\mathrm{prop}}$, $T_{\mathrm{TIA}}$, and $T_{\mathrm{ADC}}$ denote the single-sample latencies of the DAC, propagation of the optical carrier, TIA, and ADC, respectively. 
The optical modulation of a sequence of length $n$ at baud rate $f_B$ requires a duration
\begin{equation}
T_{\mathrm{mod}} = \frac{n}{f_B}.
\end{equation}

We continue analyzing the individual single-sample latencies for each stage of the system.

\subsection*{Optical Propagation}

The propagation delay through a TFLN MZM of length $L_{\mathrm{MZM}}$ is
\begin{equation}
T_{\mathrm{prop}} = \frac{n_{\mathrm{eff}} L_{\mathrm{MZM}}}{c}.
\end{equation}

For $L_{\mathrm{MZM}} = 7.3\,\mathrm{mm}$ and $n_{\mathrm{eff}} = 1.2$, this yields
\begin{equation}
T_{\mathrm{prop}} \approx 29\,\mathrm{ps}\;.
\end{equation}

\subsection*{Analog Accumulation}
The settling time of the photodetection stage is determined by the TIA bandwidth $f_{3\mathrm{dB}}$:
\begin{equation}
T_{\mathrm{TIA}} \approx \frac{5}{2\pi f_{3\mathrm{dB}}},
\end{equation}
where this conservative choice corresponds to a residual error of approximately $0.67\%$.

The first TIA of the \textit{Optmax} imposes relatively relaxed bandwidth requirements, as it only needs to accumulate the signal over the full sequence length.
The second TIA, placed before the ADC, requires a bandwidth of at least \(f_B\) to ensure accurate signal acquisition.
To remain conservative, we assume that both TIAs operate with the same minimum bandwidth, $f_{3\mathrm{dB}} \approx f_B = 40\,\mathrm{GHz}$, yielding:
\begin{equation}
T_{\mathrm{TIA}} \approx 20\,\mathrm{ps}.
\end{equation}

\subsection*{Mixed-Signal Conversion}
The lower bound for the latency of the DAC and ADC stages is given by the inverse of the sampling rate.
For our system, we consider a sampling rate of \(4f_B\) for both the ADC and DAC, corresponding to 4 samples per symbol, with \(f_B = 10\,\mathrm{GBaud}\). 
Thus, we have
\begin{equation}
T_{\mathrm{DAC}} = T_{\mathrm{ADC}} = \frac{1}{4f_B} = 25\,\mathrm{ps}.
\end{equation}

%Alternatively, we could use the estimated conversion time reported in Ref.~\cite{alkayed2025photonicising}, namely $T_{\mathrm{DAC,max}} \approx 5\,\mathrm{ns}$.

\subsection*{Total Latency}
We consider a pipelined operation, where each stage processes samples concurrently: while the $k$-th sample is being converted by the ADC, the $(k+1)$-th sample is being amplified by the TIA, the $(k+2)$-th sample is propagating through the optical domain, and subsequent samples are being encoded into the optical carrier by the DAC.
The overall throughput is therefore determined by the symbol rate $f_B$, while the total latency is given by the sum of (i) the time required to process the $n$ samples at rate $f_B$, and (ii) a constant offset corresponding to the cumulative delay of the pipeline stages.
The \textit{Optmax} is composed of two trains; the latency of the first train is given by
\begin{equation}
T_1 = \frac{n}{f_B} + (T_{\mathrm{DAC}} + T_{\mathrm{prop}}+ T_{\mathrm{TIA}})\;,
\end{equation}
while the latency of the second train is
\begin{equation}
T_2 = \frac{n}{f_B} + (T_{\mathrm{DAC}} + T_{\mathrm{prop}} + T_{\mathrm{TIA}} + T_{\mathrm{ADC}})\;.
\end{equation}
Thus, considering a sequence length $n=2048$, the total latency of the \textit{Optmax} architecture is given by
\begin{equation}
T_{\mathrm{Optmax}} = 2\frac{n}{f_B} + 2T_{\mathrm{DAC}} + 2T_{\mathrm{prop}} + 2T_{\mathrm{TIA}} + T_{\mathrm{ADC}} \approx \SI{410}{ns} \; ,
\end{equation}
while, for the \textit{Optmoid} architecture, we obtain
\begin{equation}
T_{\mathrm{Optmoid}} = \frac{n}{f_B} + T_{\mathrm{DAC}} + T_{\mathrm{prop}} + T_{\mathrm{TIA}} + T_{\mathrm{ADC}} \approx \SI{205}{ns} \; .
\end{equation}

\begin{figure*}
    \centering
    \includegraphics[width=\linewidth]{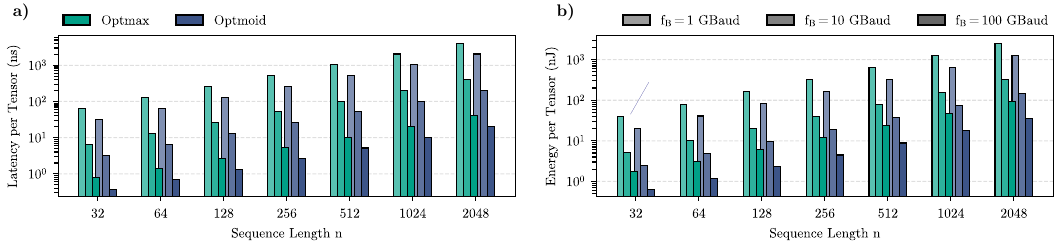}
    \caption{\textbf{Latency and compute efficiency scaling of \textit{Optmax} and \textit{Optmoid}} \textbf{(a)} Estimated latency per Softmax operation as a function of the sequence length \(N\) for three baud rates (\(f_B=1\), \(10\), and \(100\,\mathrm{GBaud}\)). The analysis follows the pipelined latency model described in Supplementary Sec.~\ref{supp:latency}, including digital-to-analog conversion, optical propagation through the TFLN MZMs, photodetection with transimpedance amplification, and analog-to-digital conversion. For all operating points, \textit{Optmoid} exhibits lower latency than \textit{Optmax} because it requires only a single modulation stage. \textbf{(b)} Estimated energy per Softmax operation as a function of the sequence length \(N\) for three baud rates (\(f_B=1\), \(10\), and \(100\,\mathrm{GBaud}\)). The energy is computed from the total electrical power consumption of each architecture, including the laser source, TFLN MZMs, DAC, RF driver, photodetection stage, and ADC, multiplied by the corresponding operation latency. \textit{Optmoid} yields a lower energy per operation than \textit{Optmax} across the full parameter range owing to its single-stage architecture.}
    \label{fig:supp_latency_and_energy}
\end{figure*}

Finally, we report the calculated latency per input of length $n$ for \textit{Optmoid} and \textit{Sigmoid} and different baudrates in \cref{fig:supp_latency_and_energy}\textcolor{blue}{a}.

\section{Power Consumption}
\label{supp:power}

\subsection*{System Overview}

As described in the main text, the \textit{Optmax} architecture consists of an MZM followed by a slow photodetector and a transimpedance amplifier (TIA) for analog accumulation, whose output drives a second MZM. The resulting signal is then detected by a fast photodiode, amplified by a second TIA, and finally converted to the digital domain by an ADC.
By contrast, the \textit{Optmoid} architecture uses a single MZM followed by a photodetector and TIA, with the output then directly digitized by an ADC.

Accordingly, the analysis is organized into the following subsystems: 
(i) the continuous-wave laser source, 
(ii) thin-film lithium niobate (TFLN) Mach--Zehnder modulators (MZMs), including the driving electronics, 
(iii) the photodetection stage (photodiodes and TIAs), and 
(iv) the analog-to-digital conversion stage (ADC). 
The total system power consumption is obtained by summing the contributions of these individual subsystems.

\subsection*{Laser Source}

The electrical power consumption of a laser diode is given by
\begin{equation}
P_{\mathrm{L}} = V_{\mathrm{L}}\left(\frac{P_{\mathrm{opt}}}{\gamma_{\mathrm{L}}} + I_{\mathrm{th}}\right),
\end{equation}
where $V_{\mathrm{L}}$ is the bias voltage, $\gamma_{\mathrm{L}}$ the slope efficiency, $I_{\mathrm{th}}$ the threshold current, and $P_{\mathrm{opt}}$ the emitted optical power.

These parameters depend on the specific laser source. Using representative values for the laser employed in this work (Toptica CTL 1550 laser, $V_{\mathrm{L}} = 1.6\,\mathrm{V}$, $\gamma_{\mathrm{L}} = 0.24\,\mathrm{mW/mA}$, $I_{\mathrm{th}} = 60\,\mathrm{mA}$, $P_{\mathrm{opt}} = 50\,\mathrm{mW}$), we obtain
\begin{equation}
P_{\mathrm{L}} \approx 0.43\,\mathrm{W}.
\end{equation}

\subsection*{Electro-Optic Modulation (TFLN MZMs)}
The power consumption of the MZMs is determined by the RF power dissipated in the $50\,\mathrm{\Omega}$ termination and by the DC power required to drive the thermal phase shifters used to maintain quadrature operation.
Additional power consumption arises from the electrical drive circuitry, consisting of a DAC and RF driver amplifier.
Note that the second MZM of the \textit{Optmax} architecture is driven by a low-frequency signal from the TIA output, so it does not require the RF driver amplifier.

\textbf{MZM.}
The power dissipated in the MZM termination and the thermal phase shifter is
\begin{equation}
  P_{\mathrm{MZM}} = \frac{V_\text{RMS}^2}{R} + P_{\mathrm{DC}} \;,
\end{equation}
where $V_\text{RMS}$ is the root-mean-square voltage required for modulation, $R=50\,\mathrm{\Omega}$ is the termination resistance, and $P_{\mathrm{DC}}$ is the power required for the thermal phase shifter.

The voltage range required for the \textit{Optmax} architecture is $V_{\mathrm{max}} \approx 2.87\,\mathrm{V}$, while for the \textit{Optmoid} architecture it is $V_{\mathrm{max}} = 5.73\,\mathrm{V}$.
Assuming uniformly distributed values of input voltages in the interval $[-V_{\mathrm{max}}/2, V_{\mathrm{max}}/2]$, the RMS voltage is given by $V_\text{RMS} = V_{\mathrm{max}}/(2\sqrt{3})$.

The power consumed by the thermal phase shifters is estimated from the electrical power delivered by the DC source. 
We operated the source (Keysight E36106A) at different bias points for \textit{Optmax} and \textit{Optmoid}. 
For \textit{Optmoid}, we biased the MZM at the quadrature point, whereas for \textit{Optmax} the operating point was set between the minimum-transmission point and the quadrature point. 
The corresponding electrical power consumption was \(P_\text{DC,Optmax}=2.2\,\text{V}\times 6.28\,\text{mA}=13.8\,\text{mW}\) for \textit{Optmax} and \(P_\text{DC,Optmoid}=3\,\text{V}\times 8.55\,\text{mA}=25.6\,\text{mW}\) for \textit{Optmoid}.

The total power consumed in the MZMs is therefore
\begin{align}
  &P_{\text{MZM}}^\text{Optmax} = 13.7\text{mW} + 13.8\,\text{mW} =  27.5\,\text{mW} \;, \\
  &P_{\text{MZM}}^\text{Optmoid} = 54.7\text{mW} + 25.6\,\text{mW} = 80.3\,\text{mW} \;.
\end{align}

\textbf{DAC.}
To estimate the DAC power consumption, we follow the approach reported in Ref. \cite{al-kayed_programmable_2025}, which assumes that DAC power scales linearly with the sampling rate~\cite{jiang2022dmtserdes}. 
In that work, the authors report a power consumption of \(P_{\mathrm{DAC}}' = 132.25\,\mathrm{mW}\) at a sampling rate of \(97\,\mathrm{GS/s}\). 
For our system, we consider a sampling rate of \(4f_B\), corresponding to 4 samples per symbol, with \(f_B = 10\,\mathrm{GBaud}\). 
We thus obtain
\begin{equation}
P_{\mathrm{DAC}} = P_{\mathrm{DAC}}' \frac{4f_B}{97 \,\mathrm{GS/s}} \approx 54.5\,\mathrm{mW} \;.
\end{equation}

\textbf{Driver amplifier.}
Following Ref. \cite{al-kayed_programmable_2025}, we estimate an integrated RF driver power of
\begin{equation}
P_{\mathrm{drive}} = 100\,\mathrm{mW}.
\end{equation}

\subsection*{Photodetection and Analog Front-End}

The photodetection stage consists of a reverse-biased photodiode and a transimpedance amplifier (TIA).

The photodiode contributes on the order of \(P_{\mathrm{PD}} \approx 1\,\mathrm{mW}\) \cite{al-kayed_programmable_2025}.
Following the same approach used for the latency calculation, we assume that both TIAs operate with identical bandwidths \(f_B\), and we therefore assign each a power consumption of \(P_{\mathrm{TIA}} \approx 11.2\,\mathrm{mW}\)~\cite{daneshgar2022pam4tia}.

\subsection*{ADC}
For the ADC, we follow the same scaling argument used for the DAC, obtaining
\[
P_\text{ADC} = P_{ADC}' \frac{4f_B}{97 \ \text{GS/s}} = 130.25 \ \text{mW} \times \frac{40}{97} \approx 53.7 \ \text{mW}
\]
where $P_{ADC}'=130.25 \ \text{mW}$ is the power reported in Ref. \cite{al-kayed_programmable_2025} at $97 \ \text{GS/s}$, while \(4f_B\) denotes the ADC sampling rate assuming 4 samples per symbol, with \(f_B=10 \ \text{GBaud}\) the modulator symbol rate.

\subsection*{Total Power Consumption}
For the \textbf{Optmax} system, the total power consumption is given by
\begin{equation}
\begin{aligned}
&P_{\mathrm{tot}}^{\mathrm{Optmax}} = \\
&P_{\mathrm{L}} + 2P_{\mathrm{MZM}}^\text{Optmax} + 2P_{\mathrm{DAC}} \\
& + P_\text{drive} + 2P_{\mathrm{PD}} + 2P_{\mathrm{TIA}} + P_{\mathrm{ADC}}\;,
\end{aligned}
\end{equation}
yielding
\begin{equation}
P_{\mathrm{tot}}^{\mathrm{Optmax}} \approx 772\,\mathrm{mW}.
\end{equation}

For the \textbf{\textit{Optmoid}} system, we have the total power consumption equal to
\begin{equation}
\begin{aligned}
&P_{\mathrm{tot}}^{\mathrm{Optmoid}} = \\
&P_{\mathrm{L}} + P_{\mathrm{MZM}}^\text{Optmoid} + P_{\mathrm{DAC}} \\
&+ P_\text{drive} + P_{\mathrm{PD}} + P_{\mathrm{TIA}} + P_{\mathrm{ADC}}\;,
\end{aligned}
\end{equation}
yielding
\begin{equation}
P_{\mathrm{tot}}^{\mathrm{Optmoid}} \approx 731\,\mathrm{mW}.
\end{equation}

\subsection*{Compute Efficiency}

For $n = 2048$ and $T$ is the total latency computed in the previous section, we can compute the number of operations per second as $n/T$, and the energy consumption per operation as $E = P_{\mathrm{tot}} / (n/T)$.

\textbf{Optmax:}
The number of \textit{Optmax} operations per second is
\begin{equation}
    \frac{n}{T} = \frac{2048}{\SI{410}{ns}} \approx 5 \times 10^{9}\ \text{Optmax/s}
\end{equation}
while the energy consumption per \textit{Optmax} operation is
\begin{equation}
    E = \frac{P_{\mathrm{tot}}^\text{Optmax}}{n/T} \approx 154\,\mathrm{pJ/Optmax}.
\end{equation}

\textbf{Optmoid:}
The number of \textit{Optmoid} operations per second is
\begin{equation}
    \frac{n}{T} = \frac{2048}{ \SI{205}{ns}} \approx 10^{10}\ \text{Optmoid/s},
\end{equation}
and the energy consumption per \textit{Optmoid} operation is
\begin{equation}
    E \approx 73.1\,\mathrm{pJ/Optmoid}.
\end{equation}

These results demonstrate that electro-optic nonlinearities enable sub-microsecond latency and sub-nJ energy per operation, competitive with state-of-the-art digital implementations.

Finally, we report the energy per input of length $n$ for \textit{Optmoid} and \textit{Sigmoid} and different baudrates in \cref{fig:supp_latency_and_energy}\textcolor{blue}{b}.

\section{Simulation Details}
\label{supp:simulations_details}

\subsection*{Image Classification}
\label{Supp_vit}
We train a ViT in the standard configuration listed in \cref{tab:vit_config} using fp32 precision on the four nonlinearities (Softmax, Sigmoid, \textit{Optmax}, and \textit{Optmoid}), swapping only the attention nonlinearity while keeping all other model- and training hyperparameters fixed.

\begin{table}[t]
    \centering
    \caption{ViT model configuration.}
    \label{tab:vit_config}
    \begin{tabular}{lc}
        \toprule
        Parameter & Value \\
        \midrule
        Embedding dimension & 256 \\
        Hidden dimension & 512 \\
        Number of heads & 8 \\
        Number of layers & 6 \\
        Patch size & 4 \\
        Number of channels & 3 \\
        Number of patches & 64 \\
        Number of classes & 10 \\
        Dropout probability & 0.2 \\
        \bottomrule
    \end{tabular}
\end{table}

As detailed in Methods, the \textit{Optmax} activation requires calibration via tuning of the $x$ and $z$ clipping ranges, while \textit{Optmoid} and Sigmoid require tuning of the bias $b$. All hyperparameters were tuned on the CIFAR-10 validation split \cite{cifar_2009} by training for 200 epochs with a batch size of 128 at full precision (no quantization). 

The results of the Softmax learning rate sweep are provided in \cref{tab:lr_sweep}; based on these, a learning rate of $5 \times 10^{-4}$ was selected for all subsequent experiments. We tune the \textit{Optmax} input clipping range $[x_{\min}, x_{\max}]$ using a vanilla reciprocal $1/z$. As shown in \cref{tab:optmax_xclip}, the maximum evaluation accuracy was achieved at $[x_{\min}, x_{\max}] = [0, 10]$. We subsequently tuned the reciprocal clipping range $[z_{\min}, z_{\max}]$, with results in \cref{tab:optmax_zclip} indicating peak performance at $[z_{\min}, z_{\max}] = [1.5, 6.5]$.

The bias $b$ for both Sigmoid and \textit{Optmoid} was tuned around the negative logarithmic sequence length \cite{ramapuram_sigmoid_2025} of $b = -\ln(64) = -4.16$. Results in \cref{tab:vit_sigmoid_optmoid_bias} show the highest evaluation accuracy at $b = -11.16$ for Sigmoid and $b = -7.16$ for \textit{Optmoid}.

These derived values were applied to all vision datasets reported in Results. However, under harsh 4-bit quantization, \textit{Optmoid} and Sigmoid failed to converge with these tuned biases, as the resulting outputs were consistently quantized to 0. Consequently, for the experiments in Results \cref{fig:vit_results}\textcolor{blue}{d--f}, we reverted to the standard bias of $b = -\ln(64) = -4.16$. While re-tuning all hyperparameters for every specific quantization level and dataset would be ideal, we omit this process here due to computational constraints.

\begin{table}[]
    \centering
    \caption{Softmax ViT validation accuracy (\%) on CIFAR-10 for different learning rates.}
    \label{tab:lr_sweep}
    \begin{tabular}{lcccc}
        \toprule
        Learning rate & 3e-4 & 5e-4 & 7e-4 & 9e-4 \\
        \midrule
        Validation accuracy (\%) & 78.10 & \textbf{79.02} & 78.10 & 75.02 \\
        \bottomrule
    \end{tabular}
\end{table}

\begin{table}[]
    \centering
    \caption{\textit{Optmax} ViT validation accuracy (\%) for different $x_{\min}$ and $x_{\max}$ settings using vanilla $1/z$ normalizations.}
    \label{tab:optmax_xclip}
    \begin{tabular}{ccc}
        \toprule
        $x_{\min}$ & $x_{\max}$ & Val. acc. (\%) \\
        \midrule
        -14 & 14 & 77.72 \\
        -12 & 12 & 77.18 \\
        -10 & 10 & 77.46 \\
        -6 & 6 & 76.52 \\
        -4 & 4 & 76.16 \\
        -2 & 2 & 75.88 \\
        0 & 2 & 77.78 \\
        0 & 4 & 78.20 \\
        0 & 6 & 78.12 \\
        0 & 8 & 79.12 \\
        0 & 10 & \textbf{79.30} \\
        0 & 12 & 78.26 \\
        0 & 14 & 78.76 \\
        0 & 16 & 78.60 \\
        0 & 18 & 78.26 \\
        \bottomrule
    \end{tabular}
\end{table}

\begin{table}[H]
    \centering
    \caption{\textit{Optmax} ViT validation accuracy (\%) for different $z_{\min}$ and $z_{\max}$ settings using $[x_{\min}, x_{\max}]=[0,10]$.}
    \label{tab:optmax_zclip}
    \begin{tabular}{ccc}
        \toprule
        $z_{\min}$ & $z_{\max}$ & Val. acc. (\%) \\
        \midrule
            0.5 & 5.5 & 79.78 \\
            0.5 & 6.5 & 79.90 \\
            0.5 & 7.5 & 79.66 \\
            0.5 & 8.5 & 79.96 \\
            0.5 & 9.5 & 79.60 \\
            0.5 & 10.5 & 79.36 \\
            1.5 & 5.5 & 79.66 \\
            1.5 & 6.5 & \textbf{80.42} \\
            1.5 & 7.5 & 79.86 \\
            1.5 & 8.5 & 80.16 \\
            1.5 & 9.5 & 79.94 \\
            1.5 & 10.5 & 79.60 \\
            2.5 & 5.5 & 80.18 \\
            2.5 & 6.5 & 79.46 \\
            2.5 & 7.5 & 80.14 \\
            2.5 & 8.5 & 79.94 \\
            2.5 & 9.5 & 79.62 \\
            2.5 & 10.5 & 78.98 \\
            3.5 & 5.5 & 78.66 \\
            3.5 & 6.5 & 79.72 \\
            3.5 & 7.5 & 78.58 \\
            3.5 & 8.5 & 79.66 \\
            3.5 & 9.5 & 79.50 \\
            3.5 & 10.5 & 79.62 \\
        \bottomrule
    \end{tabular}
\end{table}

\begin{table}[]
    \centering
    \caption{Sigmoid and \textit{Optmoid} ViT validation accuracy (\%) for bias $b$.}
    \label{tab:vit_sigmoid_optmoid_bias}
    \begin{tabular}{ccc}
        \toprule
        Bias $b$ & \makecell{Sigmoid\\Val. acc. (\%)} & \makecell{\textit{Optmoid}\\Val. acc. (\%)} \\
        \midrule
            -17.16 & 79.22 & 79.54 \\
            -16.16 & 79.24 & 79.54 \\
            -15.16 & 80.08 & 79.54 \\
            -14.16 & 80.00 & 79.06 \\
            -13.16 & 80.18 & 79.06 \\
            -12.16 & 80.04 & 79.28 \\
            -11.16 & \textbf{80.52} & 79.26 \\
            -10.16 & 79.06 & 79.14 \\
            -9.16 & 78.82 & 77.40 \\
            -8.16 & 79.32 & 78.86 \\
            -7.16 & 78.78 & \textbf{80.94}  \\
            -6.16 & 78.30 & 78.74 \\
            -5.16 & 76.92 & 77.88 \\
            -4.16 & 67.72 & 75.56 \\
            -3.16 & 66.70 & 67.78 \\
            -2.16 & 67.24 & 67.48 \\
            -1.16 & 67.82 & 67.64 \\
            -0.16 & 66.04 & 65.96 \\
        \bottomrule
    \end{tabular}
\end{table}

\subsection*{Causal Language Modeling}
\label{Supp_clm}

We train GPT-2 (124M parameters) \cite{radford_language_2019} from scratch on the FineWeb-Edu dataset \citep{lozhkov_fineweb-edu_2024}. GPT‑2 uses causal self-attention with 12 heads, a head dimension of 64, and sequences of 1,024 tokens. Softmax uses standard normalization without stretching, quantization, or noise. Optimization is performed via AdamW with a weight decay of 0.1, a learning rate of $6 \times 10^{-4}$, and a warmup-stable-decay schedule consisting of a 5\% linear warmup and a 28.5\% linear decay. Attention logits were scaled by \(1/\sqrt{64}\) and causally masked before normalization; fixed-length sequences required no padding, and masked weights were set exactly to zero. Attention-weight dropout was 0.1 during training and disabled during evaluation. Training is conducted in bf16 mixed precision—distinct from the specific input/output quantization of the attention nonlinearities—with an effective batch size of 512 sequences, yielding approximately 0.5M tokens per optimization step. For the primary comparison in the main Results (\cref{fig:clm_results}\textcolor{blue}{b}), we train four attention nonlinearities (Softmax, Sigmoid, \textit{Optmax}, and \textit{Optmoid}) for 5,000 steps ($\approx$2.6B tokens) by swapping only the nonlinearity while keeping all other model and training hyperparameters fixed. Hyperparameters were selected prior based on validation performance with training runs of fewer steps.

The dataset is partitioned deterministically into train (94\%), validation (5\%), and test (1\%) splits. Validation and test loss are evaluated every 125 steps on 50 batches, while final test metrics are reported based on an evaluation of 200 batches. For \textit{Optmax}, the input clipping range $[x_{\min}, x_{\max}]$ is tuned using a vanilla reciprocal $1/z$; as shown in \cref{tab:clm_optmax_xclip}, minimum evaluation loss occurs at $[x_{\min}, x_{\max}] = [0, 4]$. We subsequently tune the reciprocal clipping range $[z_{\min}, z_{\max}]$, which achieves maximum evaluation accuracy at $[z_{\min}, z_{\max}] = [1, 17]$ as indicated in \cref{tab:clm_optmax_zclip}. We do not tune for an even larger range, since the fitting process described in Methods of mapping the lowering sinusoidal swing to the perfect reciprocal breaks down for too large ranges. 
The bias $b$ for Sigmoid and \textit{Optmoid} is tuned around the negative logarithmic sequence length \cite{ramapuram_sigmoid_2025} of $b = -\ln(1024) = -6.93$. As reported in \cref{tab:clm_sigmoid_optmoid_bias}, the optimal evaluation results are at $b = -7.93$ for Sigmoid and $b = -5.93$ for \textit{Optmoid}. In a separate quantization ablation, we train each nonlinearity for 1,500 steps with symmetric quantization of the attention input and output to 4, 8, and 16 bits, alongside a full-precision baseline. For this ablation, validation loss is evaluated every 125 steps on 50 batches, with final test evaluation performed on the held-out 1\% split.

\begin{table}[H]
    \centering
    \caption{\textit{Optmax} GPT-2 validation loss for different $x_{\min}$ and $x_{\max}$ settings using vanilla $1/z$ normalizations.}
    \label{tab:clm_optmax_xclip}
    \begin{tabular}{ccc}
        \toprule
        $x_{\min}$ & $x_{\max}$ & Val. loss \\
        \midrule
            -14 & 14 & 5.985 \\
            -12 & 12 & 5.933 \\
            -10 & 10 & 5.928 \\
            -6 & 6 & 5.860 \\
            -4 & 4 & 5.832 \\
            -2 & 2 & 5.800 \\
            0 & 2 & 5.636 \\
            0 & 4 & \textbf{5.529} \\
            0 & 6 & 5.544 \\
            0 & 8 & 5.559 \\
            0 & 10 & 5.550 \\
        \bottomrule
    \end{tabular}
\end{table}

\begin{table}[H]
    \centering
    \caption{\textit{Optmax} GPT-2 validation loss for different $z_{\min}$ and $z_{\max}$ settings using $[x_{\min}, x_{\max}]=[0,4]$.}
    \label{tab:clm_optmax_zclip}
    \begin{tabular}{ccc}
        \toprule
        $z_{\min}$ & $z_{\max}$ & Val. loss \\
        \midrule
            11.0 & 13.0 & 5.553 \\
            10.0 & 14.0 & 5.555 \\
            9.0 & 15.0 & 5.559 \\
            8.0 & 16.0 & 5.536 \\
            7.0 & 17.0 & 5.531 \\
            6.0 & 18.0 & 5.518 \\
            5.0 & 19.0 & 5.515 \\
            4.0 & 20.0 & 5.519 \\
            8.0 & 10.0 & 4.736 \\
            7.0 & 11.0 & 4.728 \\
            6.0 & 12.0 & 4.689 \\
            5.0 & 13.0 & 4.694 \\
            4.0 & 14.0 & 4.683 \\
            3.0 & 15.0 & 4.655 \\
            2.0 & 16.0 & 4.654 \\
            1.0 & 17.0 & \textbf{4.597} \\
        \bottomrule
    \end{tabular}
\end{table}

\begin{table}[H]
    \centering
    \caption{Sigmoid and \textit{Optmoid} GPT-2 validation loss for bias $b$.}
    \label{tab:clm_sigmoid_optmoid_bias}
    \begin{tabular}{ccc}
        \toprule
        Bias $b$ & \makecell{Sigmoid\\Val. loss} & \makecell{\textit{Optmoid}\\Val. loss} \\
        \midrule
            -8.93 & 5.645 & 5.967 \\
            -7.93 & \textbf{5.575} & 5.967 \\
            -6.93 & 5.580 & 5.967 \\
            -5.93 & 5.759 & 5.967 \\
            -4.93 & 5.921 & 5.759 \\
            -3.93 & 5.903 & \textbf{5.713} \\
            -2.93 & 5.854 & 6.008 \\
            -1.93 & 5.875 & 5.968 \\
            -0.93 & 5.901 & 5.949 \\
        \bottomrule
    \end{tabular}
\end{table}

%% file: refs.bib
@STRING{ICML = {Proc. Int. Conf. on Machine Learning (ICML)} }

@STRING{ICLR = {Proc. Int. Conf. on Learning Representations (ICLR)} }

@STRING{NIPS = {Proc. Neural Information Processing Systems (NeurIPS)}}

@string{EMNLP = {Proc. Conf. on Empirical Methods in Natural Language Processing (EMNLP)}}

@inproceedings{dao2023flashattention2,
  title={Flash{A}ttention-2: Faster Attention with Better Parallelism and Work Partitioning},
  author={Dao, Tri},
  booktitle=ICLR,
  year={2024}
}

@inproceedings{dao2022flashattention,
  title={Flash{A}ttention: Fast and Memory-Efficient Exact Attention with {IO}-Awareness},
  author={Dao, Tri and Fu, Daniel Y. and Ermon, Stefano and Rudra, Atri and R{\'e}, Christopher},
  booktitle= NIPS,
  year={2022}
}

@misc{zadouri2026flashattention4algorithmkernelpipelining,
      title={FlashAttention-4: Algorithm and Kernel Pipelining Co-Design for Asymmetric Hardware Scaling}, 
      author={Ted Zadouri and Markus Hoehnerbach and Jay Shah and Timmy Liu and Vijay Thakkar and Tri Dao},
      year={2026},
      eprint={2603.05451},
      archivePrefix={arXiv},
      primaryClass={cs.CL},
      url={https://arxiv.org/abs/2603.05451}, 
}

@inproceedings{shah2024flashattention,
author = {Shah, Jay and Bikshandi, Ganesh and Zhang, Ying and Thakkar, Vijay and Ramani, Pradeep and Dao, Tri},
title = {FlashAttention-3: fast and accurate attention with asynchrony and low-precision},
year = {2024},
booktitle = NIPS,
}

@inproceedings{dosovitskiy_vit_2021,
  title     = {An Image is Worth 16x16 Words: Transformers for Image Recognition at Scale},
  author    = {Dosovitskiy, Alexey and Beyer, Lucas and Kolesnikov, Alexander and Weissenborn, Dirk and Zhai, Xiaohua and Unterthiner, Thomas and Dehghani, Mostafa and Minderer, Matthias and Heigold, Georg and Gelly, Sylvain and Uszkoreit, Jakob and Houlsby, Neil},
  booktitle = ICLR,
  year      = {2021}
}

@inproceedings{wang_vexp_2025,
	title = {{VEXP}: {A} {Low}-{Cost} {RISC}-{V} {ISA} {Extension} for {Accelerated} {Softmax} {Computation} in {Transformers}},
	shorttitle = {{VEXP}},
	author = {Wang, Run and Islamoglu, Gamze and Belano, Andrea and et al.},
	year = {2025},
    booktitle = {2025 IEEE 32nd Symposium on Computer Arithmetic (ARITH)}
}

@inproceedings{karami_understanding_2025,
	title = {Understanding the {Performance} {Horizon} of the {Latest} {ML} {Workloads} with {NonGEMM} {Workloads}},
	booktitle = {2025 {IEEE} {International} {Symposium} on {Performance} {Analysis} of {Systems} and {Software} ({ISPASS})},
	author = {Karami, Rachid and Kao, Sheng-Chun and Kwon, Hyoukjun},
	month = may,
	year = {2025},
}

@inproceedings{wang_sole_2023,
	title = {{SOLE}: {Hardware}-{Software} {Co}-design of {Softmax} and {LayerNorm} for {Efficient} {Transformer} {Inference}},
	booktitle = {2023 {IEEE}/{ACM} {International} {Conference} on {Computer} {Aided} {Design} ({ICCAD})},
	author = {Wang, Wenxun and Zhou, Shuchang and Sun, Wenyu and Sun, Peiqin and Liu, Yongpan},
	year = {2023},
}

@misc{mnist_2010,
  title={MNIST handwritten digit database},
  author={LeCun, Yann and Cortes, Corinna and Burges, Christopher J.C.},
  year={2010},
  url={http://yann.lecun.com/exdb/mnist/}
}

@inproceedings{svhn,
  author    = {Yuval Netzer and Tao Wang and Adam Coates and Alessandro Bissacco and Bo Wu and Andrew Y. Ng},
  title     = {Reading Digits in Natural Images with Unsupervised Feature Learning},
  booktitle = {NIPS Workshop on Deep Learning and Unsupervised Feature Learning},
  year      = {2011}
}

@techreport{cifar_2009,
  author = {Alex Krizhevsky},
  title = {Learning multiple layers of features from tiny images},
  institution = {University of Toronto},
  year = {2009},
  type = {Technical Report},
  url = {https://www.cs.toronto.edu/~kriz/cifar.html}
}

@inproceedings{yu_nn-lut_2022, 
author = {Yu, Joonsang and Park, Junki and Park, Seongmin and Kim, Minsoo and Lee, Sihwa and Lee, Dong Hyun and Choi, Jungwook}, 
title = {{NN-LUT}: neural approximation of non-linear operations for efficient transformer inference}, 
year = {2022}, 
isbn = {9781450391429}, 
url = {https://doi.org/10.1145/3489517.3530505}, 
doi = {10.1145/3489517.3530505}, 
booktitle = {Proceedings of the 59th ACM/IEEE Design Automation Conference}, 
}

@inproceedings{sadeghi_peano-vit_2024,
	title = {{PEANO}-{ViT}: {Power}-{Efficient} {Approximations} of {Non}-{Linearities} in {Vision} {Transformers}},
	booktitle = {Proceedings of the 29th {ACM}/{IEEE} {International} {Symposium} on {Low} {Power} {Electronics} and {Design}},
	publisher = {ACM},
	author = {Sadeghi, Mohammad Erfan and Fayyazi, Arash and Azizi, Seyedarmin and Pedram, Massoud},
	year = {2024},
}

@inproceedings{vaswani_attention_2017,
	title = {Attention is {All} you {Need}},
	booktitle = NIPS,
	author = {Vaswani, Ashish and Shazeer, Noam and Parmar, Niki and Uszkoreit, Jakob and Jones, Llion and Gomez, Aidan N and Kaiser, Ł ukasz and Polosukhin, Illia},
	year = {2017},
}

@misc{radford_language_2019,
  title={Language Models are Unsupervised Multitask Learners},
  author={Radford, Alec and Wu, Jeff and Child, Rewon and Luan, David and Amodei, Dario and Sutskever, Ilya},
  year={2019},
  url={https://cdn.openai.com/better-language-models/language_models_are_unsupervised_multitask_learners.pdf}
}

@misc{lozhkov_fineweb-edu_2024,
    author       = { Lozhkov, Anton and Ben Allal, Loubna and von Werra, Leandro and Wolf, Thomas },  
    title        = { FineWeb-Edu: the Finest Collection of Educational Content }, 
    year         = 2024,  
    url          = { https://huggingface.co/datasets/HuggingFaceFW/fineweb-edu },  
    doi          = { 10.57967/hf/2497 },
    publisher    = { Hugging Face }
}

@inproceedings{xia_hyft_2024,
	title = {Hyft: {A} {Reconfigurable} {Softmax} {Accelerator} with {Hybrid} {Numeric} {Format} for both {Training} and {Inference}},
	booktitle = {Proceedings of the 29th {ACM}/{IEEE} {International} {Symposium} on {Low} {Power} {Electronics} and {Design}},
	author = {Xia, Tianhua and Zhang, Sai Qian},
	year = {2024},
}

@inproceedings{ramapuram_sigmoid_2025,
  title={Theory, Analysis, and Best Practices for Sigmoid Self-Attention},
  author={Jason Ramapuram and Federico Danieli and Eeshan Dhekane and Floris Weers and Dan Busbridge and Pierre Ablin and Tatiana Likhomanenko and Jagrit Digani and Zijin Gu and Amitis Shidani and Russ Webb},
  booktitle=ICLR,
  year={2025},
}

@inproceedings{dash_softonic_2025,
	title = {{SOFTONIC}: {A} {Photonic} {Design} {Approach} to {Softmax} {Activation} for {High}-{Speed} {Fully} {Analog} {AI} {Acceleration}},
	booktitle = {Proceedings of the {Great} {Lakes} {Symposium} on {VLSI} 2025},
	author = {Dash, Priyabrata and Jiang, Anxiao and Dang, Dharanidhar},
	year = {2025},

}

@article{FabPaper,
title = {Redeposition-free inductively-coupled plasma etching of lithium niobate for integrated photonics},
author = {Fabian Kaufmann and Giovanni Finco and Andreas Maeder and Rachel Grange},
pages = {1601--1611},
volume = {12},
number = {8},
journal = {Nanophotonics},
doi = {doi:10.1515/nanoph-2022-0676},
year = {2023},
lastchecked = {2023-11-28}
}

@inproceedings{hoffmann2022training,
  title={Training compute-optimal large language models},
  author={Hoffmann, Jordan and Borgeaud, Sebastian and Mensch, Arthur and Buchatskaya, Elena and Cai, Trevor and Rutherford, Eliza and Casas, DDL and Hendricks, Lisa Anne and Welbl, Johannes and Clark, Aidan and others},
  booktitle=NIPS,
  year={2022}
}

@inproceedings{zhen2022cosformer,
  title={cosFormer: Rethinking Softmax In Attention},
  author={Zhen Qin and Weixuan Sun and Hui Deng and Dongxu Li and Yunshen Wei and Baohong Lv and Junjie Yan and Lingpeng Kong and Yiran Zhong},
  booktitle=ICLR,
  year={2022},
}

@article{kim2021bert,
  title={I-BERT: Integer-only BERT Quantization},
  author={Kim, Sehoon and Gholami, Amir and Yao, Zhewei and Mahoney, Michael W and Keutzer, Kurt},
  journal=ICML,
  year={2021}
}

@inproceedings{pandey_softmax_2023,
	title = {Softmax {Bias} {Correction} for {Quantized} {Generative} {Models}},
	booktitle = {2023 {IEEE}/{CVF} {International} {Conference} on {Computer} {Vision} {Workshops} ({ICCVW})},
	author = {Pandey, Nilesh Prasad and Fournarakis, Marios and Patel, Chirag and Nagel, Markus},
	year = {2023},
}

@inproceedings{Stevens_Softermax,
  author={Stevens, Jacob R. and Venkatesan, Rangharajan and Dai, Steve and Khailany, Brucek and Raghunathan, Anand},
  booktitle={2021 58th ACM/IEEE Design Automation Conference (DAC)}, 
  title={Softermax: Hardware/Software Co-Design of an Efficient Softmax for Transformers}, 
  year={2021},
  volume={},
  number={},
  pages={469-474},
  doi={10.1109/DAC18074.2021.9586134}}

@misc{park2026photonicexponentialapproximationcascaded,
      title={Photonic Exponential Approximation via Cascaded TFLN Microring Resonators toward Softmax}, 
      author={Hyoseok Park and Yeonsang Park},
      year={2026},
      eprint={2603.12934},
      archivePrefix={arXiv},
      primaryClass={physics.optics},
      url={https://arxiv.org/abs/2603.12934}, 
}

@article{sabatti_extremely_2024,
	title = {Extremely high extinction ratio electro-optic modulator via frequency upconversion to visible wavelengths},
	volume = {49},
	url = {https://opg.optica.org/ol/abstract.cfm?uri=ol-49-14-3870},
	doi = {10.1364/OL.525733},
	number = {14},
	journal = {Optics Letters},
	publisher = {Optica Publishing Group},
	author = {Sabatti, Alessandra and Kellner, Jost and Kaufmann, Fabian and Chapman, Robert J. and Finco, Giovanni and Kuttner, Tristan and Maeder, Andreas and Grange, Rachel},
	year = {2024},
	pages = {3870--3873},
}

@article{guo_deepseek-r1_2025,
	title = {{DeepSeek}-{R1} incentivizes reasoning in {LLMs} through reinforcement learning},
	volume = {645},
	doi = {10.1038/s41586-025-09422-z},
	number = {8081},
	journal = {Nature},
	author = {Guo, Daya et al},
	year = {2025},
	pages = {633--638}
}

@misc{sillman2023analogimplementationsoftmaxfunction,
      title={Analog Implementation of the Softmax Function}, 
      author={Jacob Sillman},
      year={2023},
      eprint={2305.13649},
      archivePrefix={arXiv},
      primaryClass={eess.SP},
      url={https://arxiv.org/abs/2305.13649}, 
}

@article{al-kayed_programmable_2025,
	title = {Programmable 200 {GOPS} {Hopfield}-inspired photonic {Ising} machine},
	volume = {648},
	doi = {10.1038/s41586-025-09838-7},
	language = {en},
	number = {8094},
	urldate = {2026-04-03},
	journal = {Nature},
	author = {Al-Kayed, Nayem and St-Arnault, Charles and Morison, Hugh and Aadhi, A. and Huang, Chaoran and Tait, Alexander N. and Plant, David V. and Shastri, Bhavin J.},
	month = dec,
	year = {2025},
	pages = {576--584},
}

@misc{noauthor_sdpbackend_nodate,
	title = {{SDPBackend} — {PyTorch} 2.11 documentation},
	url = {https://docs.pytorch.org/docs/stable/generated/torch.nn.attention.SDPBackend.html},
	urldate = {2026-04-04},
    year = {2026}
}

@inproceedings{paszke_pytorch_nodate,
	title = {{PyTorch}: {An} {Imperative} {Style}, {High}-{Performance} {Deep} {Learning} {Library}},
	author = {Paszke, Adam and Gross, Sam and Massa, Francisco and Lerer, Adam and Bradbury, James and Chanan, Gregory and Killeen, Trevor and Lin, Zeming and Gimelshein, Natalia},
	booktitle=NIPS,
    year={2019}
}

@misc{noauthor_firwin_nodate,
	title = {firwin — {SciPy} v1.17.0 {Manual}},
    author = {SciPy},
	url = {https://docs.scipy.org/doc/scipy/reference/generated/scipy.signal.firwin.html},
	urldate = {2026-04-07},
    year = {2026}
}

@phdthesis{jiang2022dmtserdes,
  author  = {Jiang, Z.},
  title   = {High Data Rate DMT SERDES Design},
  school  = {Carleton University},
  address = {Ottawa, Ontario},
  year    = {2022},
  doi     = {10.22215/etd/2022-14945}
}

@article{daneshgar2022pam4tia,
  author={Daneshgar, Saeid and Li, Hao and Kim, Taehwan and Balamurugan, Ganesh},
  journal={IEEE Journal of Solid-State Circuits}, 
  title={A 128 Gb/s, 11.2 mW Single-Ended PAM4 Linear TIA With 2.7 $\mu$Arms Input Noise in 22 nm FinFET CMOS}, 
  year={2022},
  volume={57},
  number={5},
  pages={1397-1408},
  doi={10.1109/JSSC.2022.3147467}}

@misc{yan2025sigmoidselfattentionlowersample,
      title={Sigmoid Self-Attention has Lower Sample Complexity than Softmax Self-Attention: A Mixture-of-Experts Perspective}, 
      author={Fanqi Yan and Huy Nguyen and Pedram Akbarian and Nhat Ho and Alessandro Rinaldo},
      year={2025},
      eprint={2502.00281},
      archivePrefix={arXiv},
      howpublished = {arXiv preprint},
      primaryClass={cs.LG},
      url={https://arxiv.org/abs/2502.00281}, 
}

@article{zhan_optoelectronic_2024,
	title = {Optoelectronic nonlinear {Softmax} operator based on diffractive neural networks},
	volume = {32},
	doi = {10.1364/OE.527843},
	number = {15},
	journal = {Optics Express},
	author = {Zhan, Ziyu and Wang, Hao and Liu, Qiang and Fu, Xing},
	year = {2024},
	pages = {26458},
}

@article{leroux_analog_2025,
	title = {Analog in-memory computing attention mechanism for fast and energy-efficient large language models},
	volume = {5},
	doi = {10.1038/s43588-025-00854-1},
	language = {en},
	number = {9},
	journal = {Nature Computational Science},
	publisher = {Nature Publishing Group},
	author = {Leroux, Nathan and Manea, Paul-Philipp and Sudarshan, Chirag and Finkbeiner, Jan and Siegel, Sebastian and Strachan, John Paul and Neftci, Emre},
	year = {2025},
	pages = {813--824},
}

@misc{kaplan2020scalinglawsneurallanguage,
      title={Scaling Laws for Neural Language Models}, 
      author={Jared Kaplan and Sam McCandlish and Tom Henighan and Tom B. Brown and Benjamin Chess and Rewon Child and Scott Gray and Alec Radford and Jeffrey Wu and Dario Amodei},
      year={2020},
      eprint={2001.08361},
      archivePrefix={arXiv},
      primaryClass={cs.LG},
      url={https://arxiv.org/abs/2001.08361}, 
      howpublished = {arXiv preprint},
}

@article{bruckerhoff-pluckelmann_case_2026,
	title = {A case study on the performance metrics of integrated photonic computing},
	volume = {4},
	issn = {2770-9019},
	url = {https://doi.org/10.1063/5.0310129},
	doi = {10.1063/5.0310129},
	number = {2},
	urldate = {2026-07-09},
	journal = {APL Machine Learning},
	author = {Brückerhoff-Plückelmann, Frank and Dijkstra, Jelle and Büchel, Julian and Batkai, Bottyan and Ebert, Falk and Mickeler, Luis and Bahr, Niklas and Egger, Urs and Sebastian, Abu and Pernice, Wolfram and Syed, Ghazi Sarwat},
	month = jun,
	year = {2026},
	pages = {026114}
}

@article{brunner2025roadmapneuromorphicphotonics,
	title = {Roadmap on {Neuromorphic} {Photonics}},
	language = {en},
	author = {Brunner, Daniel and Shastri, Bhavin J and Al-Qadasi, Mohammed A and Ballani, H and Biasi, Stefano and Bienstman, Peter and Bilodeau, Simon and Bogaerts, Wim and Böhm, Fabian and Brennan, G and Buckley, Sonia and Cai, Xinlun and Strinati, Marcello Calvanese and Canakci, B and Charbonnier, Benoit and Chemnitz, Mario and Chen, Yitong and Cheung, Stanley and Chiles, Jeff and Choi, Suyeon and Christodoulides, Demetrios N and Chrostowski, Lukas and Chu, J and Clegg, J H and Conti, Claudio and Dai, Qionghai and Lauro, Luigi Di and Dinc, Niyazi Ulas and Ewaniuk, Jacob and Fan, Shanhui and Fang, Lu and Freire, Pedro and Gentilini, Silvia and Gigan, Sylvain and Giorgi, Gian Luca and Gladrow, J and Goi, Elena and Goldmann, M and Grabulosa, A and Gu, Min and Guo, Xianxin and Hejda, Matěj and Horst, F and Hsieh, Jih-Liang and Hu, Jianqi and Hu, Juejun and Huang, Chaoran and Hurtado, Antonio and Jaurigue, Lina and Kalinin, K P and Kamalian-Kopae, Morteza and Kelly, D J and Khajavikhan, Mercedeh and Kremer, H and Laydevant, Jeremie and Lederman, Joshua C and Lee, Jongheon and Lenstra, Daan and Li, Gordon H Y and Li, Mo and Li, Yuhang and Lin, Xing and Lis, Mieszko and Lüdge, Kathy and Lugnan, Alessio and Lupo, Alessandro and Lvovsky, A I and Marandi, Alireza and Marchesin, Federico and Massar, Serge and McMahon, Peter L and Moralis-Pegios, Miltiadis and Morandotti, Roberto and Moser, Christophe and Moss, David J and Mukherjee, Avilash and Nikdast, Mahdi and Offrein, B J and Oguz, Ilker and Oripov, Bakhrom and O'Shea, G and Ozcan, Aydogan and Parmigiani, F and Pavanello, Fabio and Pavesi, Lorenzo and Peserico, Nicola and Pickup, L and Pleros, Nikos and Porte, Xavier and Primavera, Bryce A and Prucnal, Paul and Puts, Lukas and Qiao, Fei and Rahmani, B and Raineri, Fabrice and Ocampo, Carlos A Ríos and Robertson, Joshua and Romeira, Bruno and Roques-Carmes, Charles and Rotenberg, Nir and Schoenhardt, Steffen and Schwartz, Russell L T and Shainline, Jeffrey M and Skalli, A and Sohoni, Mandar M and Sorger, Volker J and Soriano, Miguel C and Stabile, Ripalta and Stiller, Birgit and Sunada, Satoshi and Tefas, Anastasios and Tsakyridis, Apostolos and Turitsyn, Sergei K and Veraldi, Daniele and Verschaffelt, Guy and Vlieg, E A and Wang, Hao and Wetzstein, Gordon and Wright, Logan G and Wu, Changming and Wu, Chu and Wu, Jiamin and Xia, Fei and Xu, Xingyuan and Yang, Hangbo and Yao, Weiming and Yildirim, Mustafa and Yoo, S J Ben and Youngblood, Nathan and Zambrini, Roberta and Zhang, Haiou and Zhang, Weipeng},  
  year={2025},
  eprint={2501.07917},
  archivePrefix={arXiv},
  primaryClass={cs.ET},
  url={https://arxiv.org/abs/2501.07917}, 
  howpublished = {arXiv preprint},
}

@inproceedings{NEURIPS2022_c1e2faff,
 author = {Hoffmann, Jordan and Borgeaud, Sebastian and Mensch, Arthur and Buchatskaya, Elena and Cai, Trevor and Rutherford, Eliza and de Las Casas, Diego and Hendricks, Lisa Anne and Welbl, Johannes and Clark, Aidan and Hennigan, Thomas and Noland, Eric and Millican, Katherine and van den Driessche, George and Damoc, Bogdan and Guy, Aurelia and Osindero, Simon and Simonyan, Kar\'{e}n and Elsen, Erich and Vinyals, Oriol and Rae, Jack and Sifre, Laurent},
 booktitle = NIPS,
 title = {An empirical analysis of compute-optimal large language model training},
 year = {2022}
}

@inproceedings{1Carion_endtoend,
author = {Carion, Nicolas and Massa, Francisco and Synnaeve, Gabriel and Usunier, Nicolas and Kirillov, Alexander and Zagoruyko, Sergey},
title = {End-to-End Object Detection with Transformers},
year = {2020},
isbn = {978-3-030-58451-1},
doi = {10.1007/978-3-030-58452-8_13},
booktitle = {Computer Vision – ECCV 2020: 16th European Conference, Glasgow, UK},
pages = {213–229},
numpages = {17},
}

@InProceedings{pmlr-v139-bertasius21a,
  title = 	 {Is Space-Time Attention All You Need for Video Understanding?},
  author =       {Bertasius, Gedas and Wang, Heng and Torresani, Lorenzo},
  booktitle = 	 ICML,
  year = 	 {2021},
}

@inproceedings{henderson-etal-2020-convert,
    title = "{C}onve{RT}: Efficient and Accurate Conversational Representations from Transformers",
    author = "Henderson, Matthew  and
      Casanueva, I{\~n}igo  and
      Mrk{\v{s}}i{\'c}, Nikola  and
      Su, Pei-Hao  and
      Wen, Tsung-Hsien  and
      Vuli{\'c}, Ivan",
    booktitle = "Findings of the Association for Computational Linguistics: EMNLP 2020",
    month = nov,
    year = "2020",
    publisher = "Association for Computational Linguistics",
    url = "https://aclanthology.org/2020.findings-emnlp.196/",
    doi = "10.18653/v1/2020.findings-emnlp.196",
    pages = "2161--2174",
}

@inproceedings{decision_transformer,
 author = {Chen, Lili and Lu, Kevin and Rajeswaran, Aravind and Lee, Kimin and Grover, Aditya and Laskin, Misha and Abbeel, Pieter and Srinivas, Aravind and Mordatch, Igor},
 booktitle = {Advances in Neural Information Processing Systems},
 pages = {15084--15097},
 publisher = {Curran Associates, Inc.},
 title = {Decision Transformer: Reinforcement Learning via Sequence Modeling},
 url = {https://proceedings.neurips.cc/paper_files/paper/2021/file/7f489f642a0ddb10272b5c31057f0663-Paper.pdf},
 volume = {34},
 year = {2021}
}

@inproceedings{rt12022arxiv,
    title={RT-1: Robotics Transformer for Real-World Control at Scale},
    author={Anthony	Brohan and  Noah Brown and  Justice Carbajal and  Yevgen Chebotar and  Joseph Dabis and  Chelsea Finn and  Keerthana Gopalakrishnan and  Karol Hausman and  Alex Herzog and  Jasmine Hsu and  Julian Ibarz and  Brian Ichter and  Alex Irpan and  Tomas Jackson and  Sally Jesmonth and  Nikhil Joshi and  Ryan Julian and  Dmitry Kalashnikov and  Yuheng Kuang and  Isabel Leal and  Kuang-Huei Lee and  Sergey Levine and  Yao Lu and  Utsav Malla and  Deeksha Manjunath and  Igor Mordatch and  Ofir Nachum and  Carolina Parada and  Jodilyn Peralta and  Emily Perez and  Karl Pertsch and  Jornell Quiambao and  Kanishka Rao and  Michael Ryoo and  Grecia Salazar and  Pannag Sanketi and  Kevin Sayed and  Jaspiar Singh and  Sumedh Sontakke and  Austin Stone and  Clayton Tan and  Huong Tran and  Vincent Vanhoucke and Steve Vega and  Quan Vuong and  Fei Xia and  Ted Xiao and  Peng Xu and  Sichun Xu and  Tianhe Yu and  Brianna Zitkovich},
    booktitle={arXiv preprint arXiv:2212.06817},
    year={2022}
}

@article{feldmann_parallel_2021,
	title = {Parallel convolutional processing using an integrated photonic tensor core},
	volume = {589},
	copyright = {2020 The Author(s), under exclusive licence to Springer Nature Limited},
	issn = {1476-4687},
	doi = {10.1038/s41586-020-03070-1},
	number = {7840},
	journal = {Nature},
	publisher = {Nature Publishing Group},
	author = {Feldmann, J. and Youngblood, N. and Karpov, M. and Gehring, H. and Li, X. and Stappers, M. and Le Gallo, M. and Fu, X. and Lukashchuk, A. and Raja, A. S. and Liu, J. and Wright, C. D. and Sebastian, A. and Kippenberg, T. J. and Pernice, W. H. P. and Bhaskaran, H.},
	month = jan,
	year = {2021},
	pages = {52--58},
}

@article{shen_deep_2017,
	title = {Deep learning with coherent nanophotonic circuits},
	volume = {11},
	issn = {1749-4885, 1749-4893},
	doi = {10.1038/nphoton.2017.93},
	language = {en},
	number = {7},
	urldate = {2026-04-13},
	journal = {Nature Photonics},
	author = {Shen, Yichen and Harris, Nicholas C. and Skirlo, Scott and Prabhu, Mihika and Baehr-Jones, Tom and Hochberg, Michael and Sun, Xin and Zhao, Shijie and Larochelle, Hugo and Englund, Dirk and Soljačić, Marin},
	month = jul,
	year = {2017},
	pages = {441--446},
}

@misc{kimiteam2026kimik2openagentic,
      title={Kimi K2: Open Agentic Intelligence}, 
      author={Kimi Team and Yifan Bai and Yiping Bao and Y. Charles and Cheng Chen and Guanduo Chen and Haiting Chen and Huarong Chen and Jiahao Chen and Ningxin Chen and Ruijue Chen and Yanru Chen and Yuankun Chen and Yutian Chen and Zhuofu Chen and Jialei Cui and Hao Ding and Mengnan Dong and Angang Du and Chenzhuang Du and Dikang Du and Yulun Du and Yu Fan and Yichen Feng and Kelin Fu and Bofei Gao and Chenxiao Gao and Hongcheng Gao and Peizhong Gao and Tong Gao and Yuyao Ge and Shangyi Geng and Qizheng Gu and Xinran Gu and Longyu Guan and Haiqing Guo and Jianhang Guo and Xiaoru Hao and Tianhong He and Weiran He and Wenyang He and Yunjia He and Chao Hong and Hao Hu and Yangyang Hu and Zhenxing Hu and Weixiao Huang and Zhiqi Huang and Zihao Huang and Tao Jiang and Zhejun Jiang and Xinyi Jin and Yongsheng Kang and Guokun Lai and Cheng Li and Fang Li and Haoyang Li and Ming Li and Wentao Li and Yang Li and Yanhao Li and Yiwei Li and Zhaowei Li and Zheming Li and Hongzhan Lin and Xiaohan Lin and Zongyu Lin and Chengyin Liu and Chenyu Liu and Hongzhang Liu and Jingyuan Liu and Junqi Liu and Liang Liu and Shaowei Liu and T. Y. Liu and Tianwei Liu and Weizhou Liu and Yangyang Liu and Yibo Liu and Yiping Liu and Yue Liu and Zhengying Liu and Enzhe Lu and Haoyu Lu and Lijun Lu and Yashuo Luo and Shengling Ma and Xinyu Ma and Yingwei Ma and Shaoguang Mao and Jie Mei and Xin Men and Yibo Miao and Siyuan Pan and Yebo Peng and Ruoyu Qin and Zeyu Qin and Bowen Qu and Zeyu Shang and Lidong Shi and Shengyuan Shi and Feifan Song and Jianlin Su and Zhengyuan Su and Lin Sui and Xinjie Sun and Flood Sung and Yunpeng Tai and Heyi Tang and Jiawen Tao and Qifeng Teng and Chaoran Tian and Chensi Wang and Dinglu Wang and Feng Wang and Hailong Wang and Haiming Wang and Jianzhou Wang and Jiaxing Wang and Jinhong Wang and Shengjie Wang and Shuyi Wang and Si Wang and Xinyuan Wang and Yao Wang and Yejie Wang and Yiqin Wang and Yuxin Wang and Yuzhi Wang and Zhaoji Wang and Zhengtao Wang and Zhengtao Wang and Zhexu Wang and Chu Wei and Qianqian Wei and Haoning Wu and Wenhao Wu and Xingzhe Wu and Yuxin Wu and Chenjun Xiao and Jin Xie and Xiaotong Xie and Weimin Xiong and Boyu Xu and Jinjing Xu and L. H. Xu and Lin Xu and Suting Xu and Weixin Xu and Xinran Xu and Yangchuan Xu and Ziyao Xu and Jing Xu and Jing Xu and Junjie Yan and Yuzi Yan and Hao Yang and Xiaofei Yang and Yi Yang and Ying Yang and Zhen Yang and Zhilin Yang and Zonghan Yang and Haotian Yao and Xingcheng Yao and Wenjie Ye and Zhuorui Ye and Bohong Yin and Longhui Yu and Enming Yuan and Hongbang Yuan and Mengjie Yuan and Siyu Yuan and Haobing Zhan and Dehao Zhang and Hao Zhang and Wanlu Zhang and Xiaobin Zhang and Yadong Zhang and Yangkun Zhang and Yichi Zhang and Yizhi Zhang and Yongting Zhang and Yu Zhang and Yutao Zhang and Yutong Zhang and Zheng Zhang and Haotian Zhao and Yikai Zhao and Zijia Zhao and Huabin Zheng and Shaojie Zheng and Longguang Zhong and Jianren Zhou and Xinyu Zhou and Zaida Zhou and Jinguo Zhu and Zhen Zhu and Weiyu Zhuang and Xinxing Zu},
      year={2026},
      eprint={2507.20534},
      archivePrefix={arXiv},
      primaryClass={cs.LG},
      url={https://arxiv.org/abs/2507.20534}, 
}

@article{lin_all-optical_2018,
	title = {All-optical machine learning using diffractive deep neural networks},
	volume = {361},
	url = {https://www.science.org/doi/full/10.1126/science.aat8084},
	doi = {10.1126/science.aat8084},
	number = {6406},
	urldate = {2026-07-09},
	journal = {Science},
	publisher = {American Association for the Advancement of Science},
	author = {Lin, Xing and Rivenson, Yair and Yardimci, Nezih T. and Veli, Muhammed and Luo, Yi and Jarrahi, Mona and Ozcan, Aydogan},
	month = sep,
	year = {2018},
}

@article{zhou_large-scale_2021,
	title = {Large-scale neuromorphic optoelectronic computing with a reconfigurable diffractive processing unit},
	volume = {15},
	copyright = {2021 The Author(s), under exclusive licence to Springer Nature Limited},
	issn = {1749-4893},
	url = {https://www.nature.com/articles/s41566-021-00796-w},
	doi = {10.1038/s41566-021-00796-w},
	language = {en},
	number = {5},
	urldate = {2026-07-09},
	journal = {Nature Photonics},
	publisher = {Nature Publishing Group},
	author = {Zhou, Tiankuang and Lin, Xing and Wu, Jiamin and Chen, Yitong and Xie, Hao and Li, Yipeng and Fan, Jingtao and Wu, Huaqiang and Fang, Lu and Dai, Qionghai},
	month = may,
	year = {2021},
	pages = {367--373},
}

@article{ahmed_universal_2025,
	title = {Universal photonic artificial intelligence acceleration},
	volume = {640},
	copyright = {2025 The Author(s), under exclusive licence to Springer Nature Limited},
	issn = {1476-4687},
	url = {https://www.nature.com/articles/s41586-025-08854-x},
	doi = {10.1038/s41586-025-08854-x},
	language = {en},
	number = {8058},
	urldate = {2026-07-09},
	journal = {Nature},
	publisher = {Nature Publishing Group},
	author = {Ahmed, Sufi R. and Baghdadi, Reza and Bernadskiy, Mikhail and Bowman, Nate and Braid, Ryan and Carr, Jim and Chen, Chen and Ciccarella, Pietro and Cole, Matthew and Cooke, John and Desai, Kishor and Dorta, Carlos and Elmhurst, Jonathan and Gardiner, Bryce and Greenwald, Elliot and Gupta, Shashank and Husbands, Parry and Jones, Brian and Kopa, Anthony and Lee, Ho John and Madhavan, Arulselvan and Mendrela, Adam and Moore, Nicholas and Nair, Lakshmi and Om, Aditya and Patel, Subie and Patro, Rutayan and Pellowski, Rob and Radhakrishnani, Esha and Sane, Sandeep and Sarkis, Nicholas and Stadolnik, Joe and Tymchenko, Mykhailo and Wang, Gongyu and Winikka, Kurt and Wleklinski, Alexandra and Zelman, Josh and Ho, Richard and Jain, Ritesh and Basumallik, Ayon and Bunandar, Darius and Harris, Nicholas C.},
	month = apr,
	year = {2025},
	pages = {368--374}
}

@article{hua_integrated_2025,
	title = {An integrated large-scale photonic accelerator with ultralow latency},
	volume = {640},
	copyright = {2025 The Author(s)},
	issn = {1476-4687},
	url = {https://www.nature.com/articles/s41586-025-08786-6},
	doi = {10.1038/s41586-025-08786-6},
	language = {en},
	number = {8058},
	urldate = {2026-08-11},
	journal = {Nature},
	publisher = {Nature Publishing Group},
	author = {Hua, Shiyue and Divita, Erwan and Yu, Shanshan and Peng, Bo and Roques-Carmes, Charles and Su, Zhan and Chen, Zhang and Bai, Yanfei and Zou, Jinghui and Zhu, Yunpeng and Xu, Yelong and Lu, Cheng-kuan and Di, Yuemiao and Chen, Hui and Jiang, Lushan and Wang, Lijie and Ou, Longwu and Zhang, Chaohong and Chen, Junjie and Zhang, Wen and Zhu, Hongyan and Kuang, Weijun and Wang, Long and Meng, Huaiyu and Steinman, Maurice and Shen, Yichen},
	month = apr,
	year = {2025},
	pages = {361--367}
}

@article{Spall:20,
author = {James Spall and Xianxin Guo and Thomas D. Barrett and A. I. Lvovsky},
journal = {Opt. Lett.},
number = {20},
pages = {5752--5755},
publisher = {Optica Publishing Group},
title = {Fully reconfigurable coherent optical vector--matrix multiplication},
volume = {45},
month = {Oct},
year = {2020},
url = {https://opg.optica.org/ol/abstract.cfm?URI=ol-45-20-5752},
doi = {10.1364/OL.401675},
}

@inproceedings{koca_hardware-efficient_2023,
	title = {Hardware-efficient {Softmax} {Approximation} for {Self}-{Attention} {Networks}},
	url = {https://ieeexplore.ieee.org/document/10181465},
	doi = {10.1109/ISCAS46773.2023.10181465},
	urldate = {2026-08-11},
	booktitle = {2023 {IEEE} {International} {Symposium} on {Circuits} and {Systems} ({ISCAS})},
	author = {Koca, Nazim Altar and Do, Anh Tuan and Chang, Chip-Hong},
	month = may,
	year = {2023},
	note = {ISSN: 2158-1525},
	pages = {1--5}
}

@article{li_high_2023,
	title = {A high speed reconfigurable architecture for softmax and {GELU} in vision transformer},
	volume = {59},
	issn = {0013-5194, 1350-911X},
	url = {https://ietresearch.onlinelibrary.wiley.com/doi/10.1049/ell2.12751},
	doi = {10.1049/ell2.12751},
	language = {en},
	number = {5},
	urldate = {2026-08-11},
	journal = {Electronics Letters},
	author = {Li, Tianyang and Zhang, Fan and Xie, Guangwei and Fan, Xitian and Gao, Yanzhao and Sun, Mingqian},
	month = mar,
	year = {2023},
	pages = {e12751}
}

@inproceedings{kim_full_2023,
	title = {Full {Stack} {Optimization} of {Transformer} {Inference}},
	url = {https://arxiv.org/abs/2302.14017},
	language = {en},
	urldate = {2026-08-11},
	author = {Kim, Sehoon and Hooper, Coleman and Wattanawong, Thanakul and Kang, Minwoo and Yan, Ruohan and Genc, Hasan and Dinh, Grace and Huang, Qijing and Keutzer, Kurt and Mahoney, Michael W. and Shao, Sophia and Gholami, Amir},
	month = jun,
    booktitle = {Architecture and System Support for Transformer Models (ASSYST), ISCA},
	year = {2023}
}

@article{bai_survey_2026,
	title = {A {Survey} of {Softmax} {Acceleration}: {From} {Algorithms} and {Function} {Realization} to {System} and {Hardware} {Optimization}},
	volume = {179},
	issn = {13837621},
	shorttitle = {A {Survey} of {Softmax} {Acceleration}},
	url = {https://linkinghub.elsevier.com/retrieve/pii/S1383762126002638},
	doi = {10.1016/j.sysarc.2026.103945},
	language = {en},
	urldate = {2026-08-11},
	journal = {Journal of Systems Architecture},
	author = {Bai, Jingyu and Fan, Jihui and Li, Hu},
	month = oct,
	year = {2026},
	pages = {103945},
}

@misc{team_kimi_2026,
	title = {Kimi {K3}: {Open} {Frontier} {Intelligence}},
	shorttitle = {Kimi {K3}},
	url = {http://arxiv.org/abs/2607.24653},
	doi = {10.48550/arXiv.2607.24653},
	urldate = {2026-08-11},
	publisher = {arXiv},
	author = {Team, Kimi and Bai, Tongtong and Bai, Yifan and Bao, Yiping and C, M. and Cai, Jianfeng and Cai, Xinyuan and Cao, Peizhou and Cao, Yuxuan and Chai, Ziwei and Charles, Y. and Che, H. S. and Chen, Guanduo and Chen, Guangyu and Chen, Guanzheng and Chen, Huarong and Chen, Jia and Chen, Jianlong and Chen, Jun and Chen, Kexin and Chen, Peng and Chen, Ruijue and Chen, Wentao and Chen, Xin and Chen, Yang and Chen, Yanru and Chen, Yifei and Chen, Yingjiang and Chen, Yuankun and Chen, Yujie and Chen, Yutian and Chen, Zhirong and Cheng, Dazhi and Cheng, Yean and Cui, Jialei and Cui, Jingbing and Dai, Anqi and Deng, Jiaqi and Ding, Hao and Ding, Rui and Ding, Shaofeng and Dong, Mengfan and Dong, Mengnan and Dong, Yuhao and Dong, Yuxin and Du, Angang and Du, Chenzhuang and Du, Dikang and Du, Jusen and Du, Yulun and Fan, Yu and Feng, Jing and Feng, Qiulin and Feng, Yichen and Fu, Kelin and Fu, Qiang and Gao, Fuxuan and Gao, Hongcheng and Gao, Jingyue and Gao, Tong and Gao, Weijia and Geng, Shangyi and Gong, Jie and Gong, Linhu and Gong, Shengao and Gong, Xiaochen and Gu, Qizheng and Gu, Yicheng and Guan, Shuhao and Guo, Haiqing and Guo, Shiqi and Guo, Xiang and Guo, Zhengyan and Hao, Beixi and Hao, Wenxin and Hao, Xiaoru and He, Dailan and He, Haotian and He, Lehan and He, Qi and He, Weiran and He, Xinran and He, Xinyi and He, Yibo and He, Yunjia and Hong, Chao and Hong, Tiange and Hu, Hao and Hu, Jiaxi and Hu, Ruikun and Hu, Weiming and Hu, Yangyang and Hu, Zhenxing and Hua, Liang and Huang, Jinbin and Huang, Ke and Huang, Ruiyuan and Huang, Siying and Huang, Weixiao and Huang, Yan and Huang, Zhengjie and Huang, Zhiqi and Hui, Yulong and Jia, Chaobo and Jiang, Yutong and Jiang, Zhejun and Jiang, Zuoyou and Jin, Wenyi and Jin, Xinyi and Jing, Yu and Kong, Huanjun and Lai, Guokun and Li, Aidi and Li, Cheng and Li, Chengyuan and Li, Cong and Li, Fang and Li, Guanyu and Li, Haoyang and Li, Jia and Li, Junxiong and Li, Lei and Li, Letian and Li, Lincan and Li, Weihong and Li, Wentao and Li, Xintong and Li, Yang and Li, Yishen and Li, Yiwei and Li, Yuxiao and Li, Zhaowei and Li, Zhaoxi and Li, Zheming and Li, Zhengxiao and Li, Zhiyuan and Lin, Jiawei and Lin, Xiaohan and Lin, Yibo and Lin, Zichao and Lin, Ziyan and Liu, Bill and Liu, Boxiao and Liu, Chuan and Liu, Liang and Liu, Shaowei and Liu, Shudong and Liu, Shuran and Liu, Tianwei and Liu, Weizhou and Liu, Yangyang and Liu, Yanming and Liu, Yibo and Liu, Yipeng and Liu, Zhengying and Liu, Zhiheng and Lu, Enzhe and Lu, Haoyu and Lu, Linqiang and Lu, Tingzhan and Lu, Zhiyuan and Luo, Aotian and Luo, G. and Luo, Junyu and Luo, Yifan and Lyu, B. and Lyu, Wenzhou and Mao, Shaoguang and Mei, Yuan and Men, Xin and Ni, Minqing and Niu, Yixuan and Pan, Siyuan and Peng, Shujun and Qi, Zhangyang and Qin, Ruoyu and Qin, ZeChao and Qin, Zeyu and Qiu, Haiquan and Qiu, Jianxin and Qiu, Jiezhong and Qu, Bowen and Qu, Yuhao and Shang, Zeyu and Shao, Youbo and Shen, Han and Shi, Jincheng and Shi, Juanfeng and Shi, Lidong and Shi, Shengyuan and Siu, Wingchun and Song, Pengwei and Song, Xiaoxi and Su, Jianlin and Su, Yunfeng and Su, Zhaochen and Sui, Lin and Sun, Jingsong and Sun, Junyao and Sun, Shaoning and Sun, Shuzhe and Sun, Tongyu and Sun, Yujun and Tai, Yunpeng and Tang, Chuning and Tang, Heyi and Tang, Sirui and Tang, Zecheng and Tian, Chaoran and Tian, Rongpeng and Tian, Yu and Tu, Wei and Wang, Chensi and Wang, Chuang and Wang, Chunjie and Wang, Dinglu and Wang, Feng and Wang, Hailong and Wang, Haiming and Wang, Hao and Wang, Hao and Wang, Huaqing and Wang, Hui and Wang, Jiayi and Wang, Jinglong and Wang, Jinhong and Wang, Jiuzheng and Wang, Linian and Wang, Shaobo and Wang, Shenzhi and Wang, Shuyi and Wang, Si and Wang, Siyuan and Wang, Tianfu and Wang, Wenjue and Wang, Xingran and Wang, Xinmei and Wang, Xinyuan and Wang, Xusheng and Wang, Yalin and Wang, Yangkun and Wang, Yao and Wang, Yaoyu and Wang, Yejie and Wang, Yiqin and Wang, Yucheng and Wang, Yuzhi and Wang, Zhaoji and Wang, Zhaowei and Wang, Zhengtao and Wang, Zhenhao and Wang, Zhongsheng and Wang, Zifan and Wei, Chu and Wei, Ming and Wei, Shouxin and Wen, Zichen and Wu, Fan and Wu, Haoning and Wu, Rucong and Wu, Wenhao and Wu, Xiaoxue and Wu, Yingcong and Wu, Yongqi and Wu, Yuxin and Wu, Zijian and Xian, Xinglang and Xiang, Chenxuan and Xiang, Yuye and Xiao, Bocheng and Xiao, Chenjun and Xiao, Xin and Xie, Jin and Xie, Xiaotong and Xie, Yifeng and Xie, Zhe and Xing, Bowei and Xiong, Yiming and Xu, Baosheng and Xu, Boyu and Xu, Jiale and Xu, Jianfan and Xu, Jing and Xu, Jinjing and Xu, L. H. and Xu, Qingtao and Xu, Shuyao and Xu, Suting and Xu, Tiantian and Xu, Tianxiang and Xu, Weixin and Xu, Xinran and Xu, Yangchuan and Xu, Ye and Xu, Yueni and Xu, Ziyao and Xue, Haonan and Yan, Junjie and Yan, Yaoyao and Yang, Fan and Yang, Guangyao and Yang, Hao and Yang, Junwei and Yang, Ruoyu and Yang, Wenjie and Yang, Xiaofei and Yang, Xinyu and Yang, Yi and Yang, Yiling and Yang, Ying and Yang, Yuchen and Yang, Zhen and Yang, Zhilin and Yang, Zian and Yang, Zuhao and Yao, Haotian and Ye, Dan and Ye, Haoran and Ye, Wenjie and Ye, Zhanbo and Yin, Bohong and Yin, Haoxiang and Yin, Xietong and Yu, Chengzhen and Yu, Haozhen and Yu, Longhui and Yu, Shengnan and Yu, Shuying and Yu, Tianxiang and Yuan, Enming and Yuan, Mengjie and Yue, Tongtian and Yue, Wei and Yue, Yang and Zha, Dunyuan and Zhan, Haobing and Zhang, B. H. and Zhang, Dehao and Zhang, Fei and Zhang, Hao and Zhang, Haoyuan and Zhang, Huanyu and Zhang, Jiapei and Zhang, Jiaxuan and Zhang, Jin and Zhang, Kaiyi and Zhang, Miaozhen and Zhang, Puqi and Zhang, Qinglei and Zhang, Rong and Zhang, Rui and Zhang, Shaoshuai and Zhang, Shiyi and Zhang, Xiaobin and Zhang, Xiaoyun and Zhang, Y. and Zhang, Yangkun and Zhang, Ye and Zhang, Yichi and Zhang, Yikun and Zhang, Yizhi and Zhang, Yongting and Zhang, Yu and Zhang, Yutao and Zhang, Yutong and Zhang, Zheng and Zhang, Zijing and Zhao, Bin and Zhao, Chenguang and Zhao, Feifan and Zhao, Jinglun and Zhao, Jinxiang and Zhao, Shuai and Zhao, Wenshuo and Zhao, Xiangyu and Zhao, Xuanle and Zhao, Yikai and Zhao, Zijia and Zheng, Haozhi and Zheng, Huabin and Zheng, Ruihan and Zheng, Shaojie and Zheng, Tengyang and Zhong, Haofeng and Zhong, Lei and Zhong, Longguang and Zhou, M. and Zhou, Qiankang and Zhou, Runjie and Zhou, Ruozhang and Zhou, Xinyu and Zhou, Yiqiao and Zhou, Zaida and Zhu, Jinguo and Zhu, Liya and Zhu, Xinhao and Zhu, Yangjunfeng and Zhu, Yuxuan and Zhu, Zhen and Zhuang, Chen and Zhuang, Weiyu and Zu, Xinxing},
	month = aug,
	year = {2026},
	note = {arXiv:2607.24653 [cs.CL]}
}

@article{PhysRevX.9.021032,
  title = {Large-Scale Optical Neural Networks Based on Photoelectric Multiplication},
  author = {Hamerly, Ryan and Bernstein, Liane and Sludds, Alexander and Solja\ifmmode \check{c}\else \v{c}\fi{}i\ifmmode \acute{c}\else \'{c}\fi{}, Marin and Englund, Dirk},
  journal = {Phys. Rev. X},
  volume = {9},
  issue = {2},
  pages = {021032},
  numpages = {12},
  year = {2019},
  month = {May},
  publisher = {American Physical Society},
  doi = {10.1103/PhysRevX.9.021032},
  url = {https://link.aps.org/doi/10.1103/PhysRevX.9.021032}
}

@INPROCEEDINGS{1010938,
  author={Zunino, R. and Gastaldo, P.},
  booktitle={2002 IEEE International Symposium on Circuits and Systems. Proceedings (Cat. No.02CH37353)}, 
  title={Analog implementation of the SoftMax function}, 
  year={2002},
  volume={2},
  number={},
  pages={II-II},
  doi={10.1109/ISCAS.2002.1010938}}

@ARTICLE{10876612,
  author={Ossieur, Peter and Moeneclaey, Bart and Coudyzer, Gertjan and Lambrecht, Joris and Craninckx, Jan and Martens, Ewout and Van Driessche, Joris and Bruynsteen, Cedric and De Busscher, Jonas and Declercq, Jakob and Gu, Ye and Niu, Shengpu and Pannier, Tinus and Singh, Nishant and Van Severen, Lucas and Bogaert, Laurens and Maes, Dennis and Vanackere, Tom and Vissers, Ewoud and Zhang, Jing and Kumari, Sulakshna and Oldenbeuving, Ruud and Zhang, Xuebing and Van Elzakker, Gijs and Van Campenhout, Joris and Absil, Philippe and Kuyken, Bart and Yin, Xin and Torfs, Guy and Missinne, Jeroen and Van Steenberge, Geert and Roelkens, Gunther and Bauwelinck, Johan},
  journal={IEEE Journal of Selected Topics in Quantum Electronics}, 
  title={Integrated Photonics and Electronics for Optical Transceivers Supporting AI/ML Applications}, 
  year={2025},
  volume={31},
  number={3: AI/ML Integrated Opto-electronics},
  pages={1-16},
  doi={10.1109/JSTQE.2025.3539379}}

@article{mcmahon_physics_2023,
	title = {The physics of optical computing},
	volume = {5},
	copyright = {2023 Springer Nature Limited},
	issn = {2522-5820},
	url = {https://www.nature.com/articles/s42254-023-00645-5},
	doi = {10.1038/s42254-023-00645-5},
	language = {en},
	number = {12},
	urldate = {2026-08-18},
	journal = {Nature Reviews Physics},
	publisher = {Nature Publishing Group},
	author = {McMahon, Peter L.},
	month = dec,
	year = {2023},
	pages = {717--734},
}

@article{kim_co-packaged_2026,
	title = {Co-packaged optics for high-performance computing and artificial intelligence},
	copyright = {2026 Springer Nature Limited},
	issn = {2520-1131},
	url = {https://www.nature.com/articles/s41928-026-01681-6},
	doi = {10.1038/s41928-026-01681-6},
	language = {en},
	urldate = {2026-08-20},
	journal = {Nature Electronics},
	publisher = {Nature Publishing Group},
	author = {Kim, Byungsoo and Choi, Soo Ho and Zograf, Georgii and Yoo, Young Jin and Baek, Yongmin and Kim, Seokho and Kim, Jongchan and Kim, Jeehwan and Chae, Sang Hoon and Kim, Hyunseok and Hong, Seunghoon and Lee, Kyusang},
	month = aug,
	year = {2026},
	pages = {1--15}
}

@ARTICLE{11358960,
  author={Yang, Shanglin and Jia, Hao},
  journal={Journal of Lightwave Technology}, 
  title={Reconfigurable Integrated Silicon Photonic Analog-to-Digital Converter}, 
  year={2026},
  volume={44},
  number={8},
  pages={2996-3003},
  doi={10.1109/JLT.2026.3655727}}

@article{song_integrated_2025,
	title = {Integrated electro-optic digital-to-analogue link for efficient computing and arbitrary waveform generation},
	volume = {19},
	copyright = {2025 The Author(s), under exclusive licence to Springer Nature Limited},
	issn = {1749-4893},
	url = {https://www.nature.com/articles/s41566-025-01719-9},
	doi = {10.1038/s41566-025-01719-9},
	language = {en},
	number = {10},
	urldate = {2026-08-20},
	journal = {Nature Photonics},
	publisher = {Nature Publishing Group},
	author = {Song, Yunxiang and Hu, Yaowen and Zhu, Xinrui and Powell, Keith and Magalhães, Letícia and Ye, Fan and Warner, Hana K. and Lu, Shengyuan and Li, Xudong and Renaud, Dylan and Lippok, Norman and Zhu, Di and Vakoc, Benjamin and Zhang, Mian and Sinclair, Neil and Lončar, Marko},
	month = oct,
	year = {2025},
	pages = {1107--1115},
}
